\documentclass{article}

\usepackage{PRIMEarxiv}

\usepackage[T1]{fontenc}
\usepackage[utf8]{inputenc}
\usepackage[english]{babel}

\usepackage{amsmath,amssymb,amsthm,bm}
\usepackage{mathtools}
\usepackage{amsfonts}
\usepackage{nicefrac}

\usepackage{microtype}

\usepackage{graphicx}
\graphicspath{{media/}}
\usepackage{booktabs,tabularx,longtable,array,multirow}
\usepackage{caption,subcaption}

\usepackage{enumitem}

\usepackage{xcolor}

\usepackage{natbib}

\usepackage{url}
\usepackage{comment}
\usepackage{lipsum}

\usepackage[hypertexnames=false]{hyperref}
\usepackage{fancyhdr}
\allowdisplaybreaks[1]

\theoremstyle{plain}
\newtheorem{theorem}{Theorem}[section]
\newtheorem{proposition}[theorem]{Proposition}
\newtheorem{lemma}[theorem]{Lemma}
\newtheorem{corollary}[theorem]{Corollary}

\theoremstyle{definition}

\newtheorem{assumption}[theorem]{Assumption}
\newtheorem{remark}[theorem]{Remark}

\newtheorem{hypothesis}[theorem]{Hypothesis}

\newcommand{\E}{\mathbb{E}}
\newcommand{\Pp}{\mathbb{P}}
\newcommand{\R}{\mathbb{R}}
\newcommand{\cF}{\mathcal{F}}
\newcommand{\cG}{\mathcal{G}}
\newcommand{\cX}{\mathcal{X}}

\newcommand{\clip}{\operatorname{clip}}

\newcommand{\Cov}{\operatorname{Cov}}

\newcommand{\eps}{\varepsilon}

\newcommand{\norm}[1]{\left\lVert #1\right\rVert}
\newcommand{\ip}[2]{\left\langle #1,#2\right\rangle}
\newcommand{\abs}[1]{\left|#1\right|}

\title{Clipped or Unclipped?
Finite-Sample Trade-offs for Averaged SGD under Heavy-Tailed Noise 
}

\author{
  Alexandra Suvorikova\\
  Independent Researcher \\
   \And
  Egor Gladin, Darina Dvinskikh \\
  HSE University \\
  \texttt{elgladin, dmdvinskikh@hse.ru} \\
  \And
  Artem Agafonov \\
  MBZUAI\\
  MIRAI\\
  \AND
  Mohammad Alkousa\\
  Innopolis University\\
  \And
  Yuriy Dorn\\
  MSU AI Institute\\
  \And
  Vladislav Matyukhin\\
  Innopolis University\\
  MIRAI\\
  \And
  Alexander Gasnikov\\
  MIRAI\\
  \texttt{gasnikov@yandex.ru} \\
}

\begin{document}
\maketitle

\begin{abstract}
Gradient clipping is widely used to stabilize training, but it need not improve the statistical accuracy of averaged SGD, even under heavy-tailed noise. We derive a finite-sample comparison of clipped and unclipped Polyak-Ruppert averaged SGD under finite conditional $p$-th moments, $p\ge2$. Our main result gives explicit accuracy and confidence conditions under which, for $p>2$, the Gaussian term dominates the unclipped deviation bound, so clipping need not improve its leading order. By balancing clipping bias and concentration, we obtain a bound in which the heavy-tail correction depends logarithmically rather than polynomially on the inverse failure probability. At $p=2$, this improves the confidence dependence of the leading bound. We establish sharpness of the unclipped heavy-tail term through an exact one-dimensional quadratic recursion and extend the comparison to projected convex SGD. We also prove concrete costs of clipping: every fixed finite threshold increases asymptotic variance on a scalar Gaussian quadratic, while whole-gradient clipping can shift the limiting point under asymmetric noise.
\end{abstract}


\section{Introduction }
Gradient clipping is widely used to control large stochastic updates and stabilize training~\cite{PascanuEtAl2013,BrownEtAl2020,touvron2023llama}. Its appeal is particularly clear under heavy-tailed noise, where rare gradients can be much larger than typical ones. For averaged stochastic gradient descent (SGD), evaluating this protection requires understanding its effect on the final estimator. Clipping suppresses extreme updates, but also changes the mean update and can attenuate the optimization signal. Its benefit therefore depends on the accuracy and confidence required from the averaged result.

We compare clipped and unclipped Polyak--Ruppert averaged SGD under finite conditional noise moments of order $p\ge2$. The comparison turns on a distinction between two contributions to the unclipped deviation bound: a Gaussian term, governed by the variance, and a heavy-tail correction, accounting for rare large increments. For $p>2$, the correction decays faster with the sample budget than the Gaussian term at fixed confidence. A threshold chosen to balance clipping bias and concentration improves the confidence dependence of the correction from polynomial to logarithmic. The consequence depends on which term controls the guarantee: once the Gaussian term dominates, clipping need not improve its leading order. At $p=2$, the confidence improvement affects the leading statistical bound itself.

Our finite-sample results make this distinction quantitative through explicit conditions on the requested accuracy and confidence. They also retain the different moment scales involved in the comparison: whole-gradient clipping acts on both signal and noise, so its bias bound depends on moments of the full stochastic gradient. Clipping only the centered noise provides an oracle benchmark for separating these effects. For projected convex optimization, the complete bounds additionally account for initialization and squared-gradient terms.

A complementary question concerns thresholds held fixed as the budget grows. Such a threshold can help at a finite horizon while changing the limiting estimator. Even when the target is preserved, a reduction in the variance of individual updates need not reduce the asymptotic variance of their averaged trajectory, because clipping also changes the local mean dynamics. We study these effects alongside the finite-sample comparison. 

\paragraph{Contributions:} Our contributions are the following:
\begin{itemize}[leftmargin=*]
\item \textbf{Finite-sample conditions for Gaussian dominance.} We derive high-probability bounds for projected convex SGD and an epoch-averaged extension for strongly convex quadratics. An accuracy-normalized boundary identifies when the unclipped heavy-tail correction is at most of Gaussian order and clarifies when clipping's confidence improvement changes the leading stochastic guarantee.

\item \textbf{Sharpness for the untrimmed averaging mechanism.} We embed a rare-outlier construction into a scalar quadratic SGD recursion, attaining the unclipped heavy-tail dependence. At $p=2$, this gives the polynomial confidence obstruction for the specified single-run averaging mechanism. Robust final aggregation and confidence amplification remain outside this lower-bound claim.

\item \textbf{Fixed-threshold effects on the target and variance.} Under the stated Polyak--Ruppert regularity conditions, every finite fixed threshold strictly increases asymptotic variance on a scalar Gaussian quadratic, while suitable thresholds can reduce it for some symmetric heavy-tailed laws. Under asymmetric noise, fixed whole-gradient clipping can shift the limiting target. For $p>2$ in the linear noise-centered setting, we give a sufficient window for thresholds growing with the run horizon to preserve the unclipped first-order limit.
\end{itemize}

An off-policy click-prediction experiment illustrates how the effect of fixed clipping can change with the budget. We leave importance weights intact, select a gradient-clipping threshold on separate source-validation runs, and hold it fixed during evaluation. The selected method improves the upper-tail target error at an intermediate budget and loses to unclipped SGD at a larger one. This provides a concrete example of a finite-horizon benefit that does not persist as the run grows.

\paragraph{Related work.}
Clipped stochastic optimization achieves logarithmic confidence dependence under weak moment assumptions \cite{GorbunovDanilovaGasnikov2020,SadievEtAl2023,NguyenEtAl2023}. Robust mean estimation and confidence amplification provide related ways to control rare deviations \cite{Catoni2012,LugosiMendelson2019,DavisEtAl2021}. Unmodified stochastic-gradient and mirror-descent methods also admit guarantees under weak moments \cite{EldowaPaudice2024,FatkhullinHueblerLan2025}. Our comparison identifies when the confidence improvement due to clipping concerns the leading stochastic term and when it concerns a lower-order correction, while keeping track of the moment scale and bias introduced by the clipping rule.

The asymptotic efficiency and finite-time concentration of averaged stochastic approximation are well studied \cite{PolyakJuditsky1992,MouEtAl2020,DurmusEtAl2025}. \cite{KoloskovaHendrikxStich2023} quantify stochastic clipping bias, and \cite{MarshallEtAl2025} analyze distribution-dependent clipping effects in high-dimensional least-squares dynamics. We connect the finite-sample accuracy--confidence comparison to the limiting target and covariance of the averaged estimator through explicit models. Together, these results explain how the budget, confidence requirement, and threshold determine the statistical trade-offs of clipping.

\paragraph{Organization.}
Section~\ref{sec:model} introduces the oracle and concentration tools. Section~\ref{sec:convex} gives the finite-sample optimization bounds; Sections~\ref{sec:asymptotics} and~\ref{sec:regimes} describe the asymptotic effects and the accuracy-normalized comparison. Section~\ref{sec:experiments} presents the off-policy experiment. The appendices provide complete proofs and further supporting results.

\section{Problem Setting and Concentration Tools}\label{sec:model}

We consider the stochastic convex optimization problem
\begin{equation}\label{eq:problem}
    \min_{x\in\cX} F(x):=\E f(x,\xi),
\end{equation}
where $\cX \subseteq \R^d$ is nonempty, closed, and convex. Assume that the solution set $\cX^\star$ is nonempty, and fix $x^\star\in\cX^\star$. Unless otherwise specified, we assume $F$ is convex, and all iterates lie in a set of diameter $D>0$,
\begin{equation}\label{eq:diameter}
    \norm{x-x^\star}\le D,
\end{equation}
where $\|\cdot\|$ is the Euclidean norm; assumption \eqref{eq:diameter} is typical for bounded feasible sets, projections onto a known ball, and stopping-time arguments (see \cite[Sec.~9.4.5]{oymak_arxiv}).

Let $\cF_{k-1}$ contain the information available before iteration $k$, for $k \ge 1$. The oracle returns
\begin{equation}
    g_k := g(x_k,\xi_k)=s_k+\zeta_k, \quad s_k\in\partial F(x_k),
\end{equation}
where $x_k$ is $\cF_{k-1}$-measurable and $\xi_k$ is fresh conditionally on $\cF_{k-1}$.

\begin{assumption}[conditional $p$-th moment of the noise $\zeta_k$]\label{ass:noise} For some $p\ge2$ and deterministic constants $\sigma_2,\sigma_p<\infty$,
\begin{equation}\label{eq:noise-moments}
    \E[\zeta_k\mid\cF_{k-1}]=0,\quad \E[\norm{\zeta_k}^2\mid\cF_{k-1}]\le\sigma_2^2,\quad \E[\norm{\zeta_k}^p\mid\cF_{k-1}]\le\sigma_p^p
\end{equation}
for every $k$. When $p=2$, set $\sigma_p=\sigma_2$.
\end{assumption}

Unless a different output is specified, the averaged iterate is $\bar x_N:=N^{-1}\sum_{k=1}^N x_k$.

\subsection{Clipping and the relevant moment scales}

For a threshold $\lambda>0$, define the clipping map  
\begin{equation}
    \operatorname{clip}_{\lambda}(y) = y\min\{1,\lambda/\|y\|\}\quad(y\ne0), \quad \operatorname{clip}_{\lambda}(0)=0.
\end{equation}
Whole-gradient clipping also requires control of the full stochastic gradient.

\begin{assumption}[moment of the full stochastic gradient]\label{ass:full-gradient} 
For constants $G_2,G_p<\infty$,
\begin{equation}\label{eq:full-gradient-moments}
    \E[\norm{g_k}^2\mid\cF_{k-1}]\le G_2^2,\quad
    \E[\norm{g_k}^p\mid\cF_{k-1}]\le G_p^p.
\end{equation}
\end{assumption}

\paragraph{Centered and full-gradient moments.} The scale $\sigma_p$ controls centered noise, whereas $G_p$ includes the predictable optimization signal. For example, if $F$ is $L$-smooth and \eqref{eq:diameter} holds, then for $x_k$ satisfying $\norm{x_k-x^\star}\le r$,
\begin{equation}\label{eq:local-Gp}
    G_p(r)\le C_p\bigl(\sigma_p+\norm{\nabla F(x^\star)}+Lr\bigr).
\end{equation}
For an unconstrained problem, or an interior optimizer, $\nabla F(x^\star)=0$; it need not vanish at a boundary solution. Thus $G_p(r)$ may exceed $\sigma_p$ if $r$ is large enough. 

\paragraph{Directions under comparison.} Our main comparison is between unclipped SGD (U) and whole-gradient clipping (WG). Noise-centered clipping (NC) serves as an oracle benchmark: 
\begin{equation}\label{eq:model-policies}
    h_k^{\rm U}=g_k,\qquad h_k^{\rm WG}=\clip_\lambda(g_k), \quad h_k^{\rm NC}=s_k+\clip_\lambda(\zeta_k)
\end{equation}
The U and WG directions use only the observed stochastic gradient. NC requires additional information because $s_k$ and $\zeta_k$ are not separately observable from $g_k$. 

The smooth quadratic extension also considers difference clipping (DC). It uses an anchor $\widetilde x$ and an estimate $\widetilde s\approx\nabla F(\widetilde x)$:
\begin{equation}\label{eq:model-policies-ctd}
    h_k^{\rm DC} := \widetilde s+ \clip_{\lambda_k}\left(g(x_k,\xi_k)-g(\widetilde x,\xi_k) \right). 
\end{equation}
The anchor and its estimate are available before drawing $\xi_k$, and both gradients use the same sample. Constructions of this type appear in \cite{GorbunovEtAl2024}. Our guarantees for DC require the anchor and difference-moment conditions in Proposition~\ref{prop:dc-certificate}.

\subsection{Fuk--Nagaev decomposition and clipping}
\label{sec:model-concentration}

The comparison separates variance-controlled fluctuations from rare large increments. For $\delta\in(0,1)$, write $\ell_\delta:=\log(4/\delta)$. Below, $C$ denotes a numerical constant and $C_p$ a constant depending only on $p$; their values may change between displays. 

\paragraph{Unclipped averages.}
Under Assumption~\ref{ass:noise} with $p>2$, the martingale Fuk--Nagaev inequality gives, with probability at least $1-\delta$,
\begin{equation}\label{eq:setting-unclipped-mean}
    \norm{\frac1N\sum_{k=1}^N\zeta_k} \le C\sigma_2\sqrt{\frac{\ell_\delta}{N}} +C_p\sigma_p\delta^{-1/p}N^{-(1-1/p)}.
\end{equation}
The general statement and its justification are given in Theorem~\ref{thm:fuk-nagaev} in Appendix~\ref{app:concentration}.

The first term has a Gaussian dependence on the confidence level and describes the accumulation of moderate increments. The second captures the contribution of rare large increments under a finite $p$-th moment. For fixed $\delta$ and $p>2$, the second term is lower order than $N^{-1/2}$. Proposition~\ref{prop:gaussian-boundary} gives the boundary at which it becomes no larger than the Gaussian term.

\paragraph{Clipped averages.}
Now use a common deterministic threshold $\lambda_k=\lambda$. For either NC or WG, let
\[
    M_p:=
    \begin{cases}
        \sigma_p, & \text{for NC},\\
        G_p, & \text{for WG}.
    \end{cases}
\]
NC uses Assumption~\ref{ass:noise}; WG additionally uses Assumption~\ref{ass:full-gradient}. For each of these choices, Proposition~\ref{prop:clipped-average} yields, with probability at least $1-\delta$,
\begin{equation}
    \norm{\frac1N\sum_{k=1}^N(h_k-s_k)} \le C \sigma_2\sqrt{\frac{\ell_\delta}{N}} + C \lambda\frac{\ell_\delta}{N} +\frac{M_p^p}{\lambda^{p-1}}.
\end{equation}
The three terms correspond to variance, bounded increments, and clipping bias, respectively. Nonexpansiveness of clipping ensures that the conditional variance of the centered clipped direction is at most $\sigma_2^2$, including for WG. The bias bound, however, depends on the moments of the quantity being clipped; see Lemma~\ref{lem:clipped-mean}.

For $M_p>0$, choosing $\lambda = M_p(N/\ell_\delta)^{1/p}$ balances the last two terms up to numerical constants; we call it the \emph{balanced threshold}. It gives
\begin{equation}\label{eq:setting-clipped-tuned}
    \norm{\frac1N\sum_{k=1}^N(h_k-s_k)} \le C\sigma_2\sqrt{\frac{\ell_\delta}{N}} +CM_p\left(\frac{\ell_\delta}{N}\right)^{1-1/p}
\end{equation}
with probability at least $1-\delta$.

Equations~\eqref{eq:setting-unclipped-mean} and~\eqref{eq:setting-clipped-tuned} exhibit the same variance-controlled scale. Their corrections differ in their confidence dependence: the unclipped correction contains $\delta^{-1/p}$, whereas the tuned clipped correction depends on $\delta$ through $\ell_\delta$. For NC, both corrections involve the same noise scale $\sigma_p$. For WG, the clipped correction involves $G_p$, which also accounts for the signal. The optimization bounds use the same concentration arguments with predictable weights or fixed linear maps. Sections~\ref{sec:convex}--\ref{sec:regimes} track the resulting moment scales and confidence dependence.

\paragraph{The endpoint $p=2$.}
Under a second-moment assumption alone, Chebyshev's inequality gives $\norm{\frac1N\sum_{k=1}^N\zeta_k} \le \frac{\sigma_2}{\sqrt{N\delta}}$ with probability at least $1-\delta$. In contrast, \eqref{eq:setting-clipped-tuned} for NC gives $C\sigma_2\sqrt{\ell_\delta/N}$. The rare-outlier construction in Proposition~\ref{prop:one-outlier-lower} shows why the $\delta^{-1/2}$ dependence cannot be uniformly removed from the ordinary empirical mean under a variance bound alone.

\section{Finite-Sample Optimization Guarantees}
\label{sec:convex}

We compare projected U-SGD and whole-gradient Clip-SGD on a convex, possibly nonsmooth objective. Fix an initial point $x_1\in\cX$. For a constant stepsize $\eta>0$ (and a constant clipping threshold $\lambda>0$ for Clip-SGD), both methods follow the recursion
\begin{equation}\label{eq:projected-update}
    x_{k+1}=\Pi_{\cX}(x_k-\eta h_k), \quad k=1,\ldots,N,
\end{equation}
where $\Pi_{\cX}$ is the Euclidean projection. We consider U-SGD with $h_k:=h_k^{\rm U}=g_k$ and whole-gradient Clip-SGD with $h_k:=h_k^{\rm WG}=\clip_\lambda(g_k)$; see~\eqref{eq:model-policies}. For the projected convex results below, Assumptions~\ref{ass:noise} and~\ref{ass:full-gradient} hold, $F$ is convex, and \eqref{eq:diameter} is satisfied. For U-SGD, only the second-moment part of Assumption~\ref{ass:full-gradient} is used. Let $R_0:=\norm{x_1-x^\star}$.

Both methods satisfy a common deterministic inequality. Since $x^\star\in\cX$ and $\Pi_\cX$ is nonexpansive, convexity and Jensen's inequality give
\begin{equation}\label{eq:convex-master}
    F(\bar x_N)-F(x^\star)
    \le \frac{R_0^2}{2\eta N} +\frac{\eta}{2N} \sum_{k=1}^N \norm{h_k}^2 +\frac1N \sum_{k=1}^N \langle h_k-s_k,\,x^\star-x_k\rangle.
\end{equation}
The last term is a weighted analogue of the averages in Section~\ref{sec:model-concentration}: the weights $x^\star-x_k$ are $\cF_{k-1}$-measurable and bounded by $D$. The martingale arguments underlying \eqref{eq:setting-unclipped-mean} and \eqref{eq:setting-clipped-tuned} therefore apply to the corresponding scalar increments, giving the same bounds with an additional factor $D$.

The squared-gradient term produces an additional remainder. For U-SGD, write $\norm{g_k}^2=\norm{s_k}^2 +2\langle s_k,\zeta_k\rangle+\norm{\zeta_k}^2$, where $\norm{s_k}^2 +\E[\norm{\zeta_k}^2\mid\cF_{k-1}] =\E[\norm{g_k}^2\mid\cF_{k-1}] \le G_2^2.$ In particular, $\norm{s_k}\le G_2$. The cross term is again a predictably weighted sum of $\zeta_k$. For the stepsize below, its contribution is absorbed into the Gaussian and heavy-tail terms using $R_0\le D$. The centered variable
$\norm{\zeta_k}^2-\E[\norm{\zeta_k}^2\mid\cF_{k 1}]$ has conditional moment of order $p/2$ bounded by $C_p\sigma_p^p$. Its concentration gives the remainder
\begin{equation}
    \mathcal Q_{p,N}(\delta):=
    \begin{cases}
    C_p\sigma_p^2 \delta^{-2/p}N^{-(1-2/p)},
    & 2<p\le4,\\
    C\sigma_p^2 \sqrt{\frac{\ell_\delta}{N}} +C_p\sigma_p^2 \delta^{-2/p}N^{-(1-2/p)},
    & p>4.
    \end{cases}
\end{equation}
The two cases use von Bahr--Esseen and Fuk--Nagaev concentration, respectively. The latter applies at moment order $p/2$, using $\E[\norm{\zeta_k}^4\mid\cF_{k-1}]\le\sigma_p^4$. For WG, $\norm{h_k}\le\lambda$, and Bernstein--Freedman concentration bounds the squared-gradient term.

The calibrated guarantees follow. Appendix~\ref{app:convex-proofs} gives the arbitrary-parameter bounds and complete proofs.

\begin{theorem}[U-SGD on a convex problem] \label{thm:convex-unclipped}
Let $p>2$ and consider U-SGD defined by \eqref{eq:projected-update} with $h_k=g_k$ and $\eta=R_0/(G_2\sqrt N)$. Then, with probability at least $1-\delta$,
\begin{equation}\label{eq:convex-unclipped-tuned}
    F(\bar x_N) - F(x^\star) \le{} \frac{R_0G_2}{\sqrt N} + C D\sigma_2\sqrt{\frac{\ell_\delta}{N}} + \frac{C_pD\sigma_p}{\delta^{1/p}N^{1-1/p}} + \frac{R_0}{2G_2\sqrt N}\mathcal Q_{p,N}(\delta).
\end{equation}
\end{theorem}

For fixed $\delta$, the final term in \eqref{eq:convex-unclipped-tuned} scales as $N^{-(3/2-2/p)}$ for $2<p\le4$; for $p>4$, its two components scale as $N^{-1}$ and $N^{-(3/2-2/p)}$. Its polynomial component is lower order than the preceding heavy-tail correction for fixed $\delta$, but not uniformly as $\delta$ decreases.

\begin{theorem}[Whole-gradient Clip-SGD on a convex problem] \label{thm:convex-clipped}
Let $p>2$ and consider whole-gradient Clip-SGD defined by \eqref{eq:projected-update} with $h_k=\clip_\lambda(g_k)$, $\eta=R_0/(G_2\sqrt N)$, and $\lambda=c_pG_p(N/\ell_\delta)^{1/p}$. Then, with probability at least $1-\delta$,
\begin{equation}\label{eq:convex-clipped-tuned}
    F(\bar x_N) - F(x^\star) \le{} \frac{R_0G_2}{\sqrt N} + C D\sigma_2\sqrt{\frac{\ell_\delta}{N}}+C_p D G_p\left(\frac{\ell_\delta}{N}\right)^{1-1/p} +\frac{C_p R_0 G_p^2}{G_2\sqrt N} \left(\frac{\ell_\delta}{N}\right)^{1-2/p}.
\end{equation}
\end{theorem}

In \eqref{eq:convex-clipped-tuned}, the squared-gradient term proportional to $R_0G_p\ell_\delta^{1/2-1/p}N^{-(1-1/p)}$ has been absorbed into the displayed $DG_p$ correction using $R_0\le D$ and $\ell_\delta\ge1$; the remaining  term is lower order in $N$ for every $p>2$ at fixed confidence.

For NC~\eqref{eq:model-policies}, the martingale and clipping-bias calculations replace $G_p$ by $\sigma_p$ and do not truncate the full predictable signal. DC~\eqref{eq:model-policies-ctd} uses a certified difference scale when its anchor error and difference moments are controlled; the corresponding quadratic guarantees are described below. Difference-clipped constructions of this kind are developed in~\cite{GorbunovEtAl2024}.

\subsection{Second-moment calibrations}

If $p=2$, we use a confidence-dependent step for U-SGD:
\begin{equation}\label{eq:convex-p2-step}
    \eta=\frac{R_0\sqrt\delta}{G_2\sqrt N}.
\end{equation}
Markov's inequality for the squared-gradient average and Chebyshev's inequality for the martingale term give
\begin{equation}\label{eq:convex-p2-optimal}
    \Pp\left( F(\bar x_N)-F(x^\star) > C (R_0 G_2 + D\sigma_2)\delta^{-1/2}N^{-1/2} \right) \le \delta.
\end{equation}
See Appendix~\ref{proof_convex-p2-optimal} for the proof.

For whole-gradient clipping, the same argument with $p=2$ and $\eta=R_0/(G_2\sqrt N)$, $\lambda=cG_2(N/\ell_\delta)^{1/2}$ gives
\begin{equation}\label{eq:convex-p2-clipped}
    F(\bar x_N) - F(x^\star) \le C R_0 G_2N^{-1/2} + C D G_2 \sqrt{\frac{\ell_\delta}{N}}
\end{equation}
with probability at least $1-\delta$. This second-moment calibration is also valid for $p>2$, but its clipping correction retains the slower $N^{-1/2}$ rate.

\subsection{Comparison and sample requirements}

For $p>2$, the tuned bounds \eqref{eq:convex-unclipped-tuned} and \eqref{eq:convex-clipped-tuned} share the optimization and Gaussian terms $R_0G_2N^{-1/2} + C D \sigma_2\sqrt{\frac{\ell_\delta}{N}}.$ Their $N^{-(1-1/p)}$ corrections differ in moment scale ($\sigma_p$ versus $G_p$) and confidence dependence ($\delta^{-1/p}$ versus $\ell_\delta^{1-1/p}$).

To compare these corrections, we first require that the squared-gradient remainders can be absorbed into them. Sufficient conditions are
\begin{equation}\label{eq:convex-remainder-regime}
    N^{1 / 2 - 1/p} \gtrsim_p \frac{R_0}{D G_2} \max\left\{\sigma_p\delta^{-1/p}, G_p \ell_\delta^{-1/p} \right\}, \quad
    N^{1/p} \gtrsim_p \frac{R_0 \sigma_p}{D G_2} \delta^{1/p} \ell_\delta^{1/2}, \quad p > 4.
\end{equation}
From now on, we write  $\gtrsim_z$ (resp. $\lesssim_z$) to hide the multiplicative constants depending on $z$. The second condition is needed only for the Gaussian component of $\mathcal Q_{p,N}$ when $p>4$.

Under \eqref{eq:convex-remainder-regime}, the U-SGD correction is no larger in order when
\begin{equation}\label{eq:convex-usgd-better}
    \sigma_p\delta^{-1/p} \lesssim_p G_p\ell_\delta^{1-1/p}, 
    \quad \text{equivalently} \quad 
    \left(\frac{\sigma_p}{G_p}\right)^p \lesssim_p\delta\ell_\delta^{p-1}.
\end{equation}
This compares the correction scales in the upper bounds.

Define the shared sample requirement $   N_0(\eps,\delta) = 1+\left(\frac{R_0G_2}{\eps}\right)^2 +\left(\frac{D\sigma_2}{\eps}\right)^2\ell_\delta.$ Since a finite sum and the corresponding maximum agree up to a numerical factor, sufficient sample sizes satisfy
\begin{equation}\label{eq:convex-complexities}
    N_{\rm U} \lesssim_p N_0(\eps,\delta) +\left(\frac{D\sigma_p}{\eps}\right)^{p/(p-1)} \delta^{-1/(p-1)},\quad 
    N_{\rm C} \lesssim_p N_0(\eps,\delta) +\left(\frac{DG_p}{\eps}\right)^{p/(p-1)} \ell_\delta,
\end{equation}
provided that horizons of these respective orders satisfy \eqref{eq:convex-remainder-regime}. Outside this regime, the sample requirements must retain the squared-gradient contributions from \eqref{eq:convex-unclipped-tuned} and \eqref{eq:convex-clipped-tuned}.

Up to constants depending only on $p$, condition~\eqref{eq:convex-usgd-better} also characterizes when the additional U-SGD sample requirement in \eqref{eq:convex-complexities} is no larger.

At $p=2$, \eqref{eq:convex-p2-optimal} and \eqref{eq:convex-p2-clipped} give
\begin{equation}\label{eq:convex-p2-complexities}
    N_{\rm U}^{(2)} \lesssim 1+\left(\frac{R_0G_2+D\sigma_2}{\eps}\right)^2\delta^{-1},\quad N_{\rm C}^{(2)} \lesssim 1+\left(\frac{R_0G_2}{\eps}\right)^2 + \left(\frac{DG_2}{\eps}\right)^2\ell_\delta.
\end{equation}
We write  $\gtrsim$ (resp. $\lesssim$) to hide the multiplicative constants. 
Polynomial confidence dependence also appears in the exact quadratic Polyak--Ruppert recursion in Proposition~\ref{cor:quadratic-usgd-all-p-lower}. This does not rule out improving confidence by running several independent copies and combining their outputs.

\subsection{Extension to strongly convex quadratics}

For unconstrained strongly convex quadratics, we analyze SGD under conditional $p$-th moment bounds on the noise, $p\ge2$. The method uses constant-step epochs, averages the second half of each epoch, and restarts from this average. Appendix~\ref{app:quadratic-epochs} gives the setup, algorithm, and proofs.

The bounds separate deterministic contraction, a Gaussian variance term, and a heavy-tail correction (Lemma~\ref{lem:one-epoch}). For fixed $p>2$ and fixed confidence, the heavy-tail correction is lower order as the target error decreases. Consequently, ordinary SGD and the noise-centered clipping benchmark have the same leading accuracy dependence in our bounds (Theorem~\ref{thm:strong-complexities}). Clipping improves the confidence dependence of the heavy-tail correction from polynomial to logarithmic. At $p=2$, this improvement affects the leading statistical term itself (Corollary~\ref{cor:quad-restarts-p2}).

The clipped guarantees extend to inputs satisfying Assumption~\ref{ass:certified-centered-input}. Difference clipping is covered when its anchor error and difference moments satisfy Proposition~\ref{prop:dc-certificate}, with the additional oracle work accounted for in \eqref{eq:dc-total-budget}.

\section{Polyak--Ruppert Asymptotics}
\label{sec:asymptotics}

The finite-sample bounds allow the threshold to depend on the horizon and confidence. We now compare the limiting target and covariance when a finite threshold is held fixed as the horizon grows, before considering growing thresholds. Complete assumptions, statements, and proofs appear in Appendix~\ref{app:strong-proofs}.

\paragraph{Unclipped SGD.}
Consider quadratic SGD with Hessian $H=H^\top\succ0$, and step sizes $\eta_k=a(k+k_0)^{-\gamma}$, $\gamma\in(1/2,1)$. Under the moment-decay condition in Proposition~\ref{prop:exact-pr} and the martingale-CLT conditions \eqref{eq:martingale-conditional-covariance}--\eqref{eq:martingale-lindeberg},
\[
    \sqrt N(\bar x_N-x^\star) \Longrightarrow \mathcal N(0,H^{-1}\Sigma H^{-1}).
\]
The exact decomposition in Proposition~\ref{prop:exact-pr} explains this limit: to first order, the averaged error is the empirical mean of the gradient noise transformed by $-H^{-1}$.

\paragraph{Fixed-threshold clipping.}
Let $h_\lambda(x)=\mathbb E[\clip_\lambda(g(x,\xi))]$ and let $x_\lambda$ be its stable root. Write $A_\lambda=\nabla h_\lambda(x_\lambda)$ and $\Sigma_\lambda=\Cov(\clip_\lambda(g(x_\lambda,\xi)))$. Under Proposition~\ref{ass:fixed-clip-pr}, Proposition~\ref{thm:fixed-clip-clt} gives
\[
    \sqrt N(\bar x_N-x_\lambda) \Longrightarrow \mathcal N(0,A_\lambda^{-1}\Sigma_\lambda A_\lambda^{-\top}).
\]
Thus clipping can change both the target and the limiting covariance. In particular, $x_\lambda\ne x^\star$ produces a persistent bias; see Proposition~\ref{prop:asymmetric-bias}.

For the scalar model $g(x,\xi)=\mu x+\xi$ with continuous symmetric noise, the target is preserved. Under the same PR regularity conditions, Proposition~\ref{cor:symmetric-variance} yields $  V_\lambda =\frac{\mathbb E\min\{\xi^2,\lambda^2\}} {\mu^2\mathbb P(|\xi|<\lambda)^2}.$ Clipping reduces the noise variance but also weakens the mean-field slope. Every finite threshold strictly increases $V_\lambda$ relative to unclipped averaging for nondegenerate Gaussian noise, whereas suitable finite thresholds can reduce it for some symmetric heavy-tailed distributions.

\paragraph{Growing thresholds.}
In the linear additive-noise model, noise-centered clipping can recover the unclipped limit while controlling the clipping bias. For i.i.d.\ noise with a finite $p$-th moment, $p>2$, and bounded or polylogarithmic $\ell_N=\log(4/\delta_N)$, Proposition~\ref{thm:growing-window} gives the sufficient choice
\[
    \lambda_N=N^\beta, \quad \frac{1}{2(p-1)}<\beta<\frac12.
\]
This makes both the clipping bias and the bounded-increment concentration term negligible on the $N^{-1/2}$ scale and preserves the covariance $H^{-1}\Sigma H^{-1}$. The accompanying finite-horizon bound, including the PR remainder, is given in \eqref{eq:growing-finite-bound}. Here one threshold $\lambda_N$ is used throughout the length-$N$ run; online thresholds and whole-gradient clipping require separate analysis.

\section{When Does Clipping Help?}
\label{sec:regimes}

We now express the stochastic comparison in terms of the requested accuracy and confidence. For $\sigma_2>0$, define the normalized accuracies
\[
    \rho_{\rm cvx}=\frac{\eps}{D\sigma_2},
    \quad
    \rho_{\rm sc}=\frac{\sqrt{\mu\eps}}{\sigma_2},
    \quad
    \chi_p=\frac{\sigma_p}{\sigma_2},
\]
for the convex and strongly convex quadratic settings, respectively. We compare the Gaussian and heavy-tail components of the bounds; the remaining optimization terms require separate control.

\begin{proposition}[Accuracy-normalized comparison]
\label{prop:accuracy-boundary}
Let $p>2$ and let $\rho$ denote the appropriate normalized accuracy. At $N\asymp\rho^{-2}\ell_\delta$, where $\ell_\delta=\log(4/\delta)$, the respective heavy-tail corrections are at most of the Gaussian order under
\begin{align}
    & \chi_p^p\rho^{p-2} \lesssim_p \delta\ell_\delta^{p-1} \qquad \text{for U-SGD}, \label{eq:accuracy-U-boundary}
    \\& \chi_p^p\rho^{p-2} \lesssim_p 1 \qquad \quad \;\;\, \text{for NC}. \label{eq:accuracy-C-boundary}
\end{align}
For whole-gradient clipping, replace $\chi_p$ in \eqref{eq:accuracy-C-boundary} by $G_p/\sigma_2$. The derivation is given in Appendix~\ref{app:crossover-algebra}. 
\end{proposition}

For fixed $p>2$ and confidence, requiring a smaller target error drives $\chi_p^p\rho^{p-2}$ to zero. Thus the bounds eventually share a Gaussian leading order, and balanced clipping changes a lower-order correction. At $p=2$, its confidence improvement affects the leading statistical term. Changes in the limiting target and covariance are addressed in Section~\ref{sec:asymptotics}. Table~\ref{tab:decision} in Appendix~\ref{app:crossover-algebra} summarizes the resulting regimes and their interpretation.

\section{Off-Policy Click Prediction}
\label{sec:experiments}
\newcommand{\OBDSharedStep}{100000}
\newcommand{\OBDSensitivitySharedStep}{300000}
\newcommand{\OBDEvaluationRuns}{512}
\newcommand{\OBDFinalHorizon}{4194304}
\newcommand{\OBDThresholdValidationRuns}{256}
\newcommand{\OBDThresholdValidationHorizon}{2097152}
\newcommand{\OBDSelectedThreshold}{5}
\newcommand{\OBDDirectFinalHorizon}{33554432}
\newcommand{\OBDDirectFinalRatio}{1.89}
\newcommand{\OBDDirectFinalRatioLow}{1.81}
\newcommand{\OBDDirectFinalRatioHigh}{1.95}
\newcommand{\OBDDirectCrossover}{8388608}
\newcommand{\OBDSelectedTuneRatio}{0.63}
\newcommand{\OBDSelectedTuneRatioLow}{0.60}
\newcommand{\OBDSelectedTuneRatioHigh}{0.65}
\newcommand{\OBDSelectedSweepRatio}{0.79}
\newcommand{\OBDSweepBestThreshold}{10}
\newcommand{\OBDSweepBestRatio}{0.67}
\newcommand{\OBDSweepBestRatioLow}{0.64}
\newcommand{\OBDSweepBestRatioHigh}{0.70}
\newcommand{\OBDEffectiveSamplePercent}{0.16\%}
\newcommand{\OBDMaxImportanceWeight}{12500}
\newcommand{\OBDPropensityFloorRows}{48}
\newcommand{\OBDFinalRatioTwoTenths}{8.40}
\newcommand{\OBDSourceFinalRatioTwoTenths}{5.83}
\newcommand{\OBDPositiveClipPercentTwoTenths}{42.8\%}
\newcommand{\OBDCrossoverTwoTenths}{262144}
\newcommand{\OBDSourceCrossoverTwoTenths}{524288}
\newcommand{\OBDSensitivityRatioTwoTenths}{8.16}
\newcommand{\OBDSensitivityCILowTwoTenths}{7.97}
\newcommand{\OBDSensitivityCIHighTwoTenths}{8.81}
\newcommand{\OBDFinalRatioHalf}{5.37}
\newcommand{\OBDSourceFinalRatioHalf}{3.77}
\newcommand{\OBDPositiveClipPercentHalf}{20.9\%}
\newcommand{\OBDCrossoverHalf}{524288}
\newcommand{\OBDSourceCrossoverHalf}{1048576}
\newcommand{\OBDSensitivityRatioHalf}{4.68}
\newcommand{\OBDSensitivityCILowHalf}{4.56}
\newcommand{\OBDSensitivityCIHighHalf}{5.05}
\newcommand{\OBDFinalRatioOne}{3.33}
\newcommand{\OBDSourceFinalRatioOne}{2.39}
\newcommand{\OBDPositiveClipPercentOne}{11.5\%}
\newcommand{\OBDCrossoverOne}{1048576}
\newcommand{\OBDSourceCrossoverOne}{2097152}
\newcommand{\OBDSensitivityRatioOne}{2.48}
\newcommand{\OBDSensitivityCILowOne}{2.41}
\newcommand{\OBDSensitivityCIHighOne}{2.68}

We test whether a threshold selected at one budget remains beneficial as the optimization horizon grows. Off-policy learning makes clipping particularly appealing: a small logging propensity produces a large importance-weighted gradient. Truncating large importance weights is a classical variance-control device in importance sampling and counterfactual learning \cite{Ionides2008,BottouEtAl2013}. Here we keep the importance weights intact and clip the stochastic gradient.

We use the full \texttt{all} campaign of the Open Bandit Dataset (OBD)~\cite{SaitoEtAl2021OBD}. The behavior policy is Bernoulli Thompson sampling (BTS), and the target policy is Uniform Random over $80$ items. If $\mu_i$ is the released probability that BTS displayed item $i$, inverse propensity scoring (IPS) reweights that observation by the target-to-logging probability ratio $\rho_i=(1/80)/\mu_i=(80\mu_i)^{-1}$. The experiment uses $12{,}357{,}200$ BTS impressions and a separately collected Random-policy log of $1{,}374{,}327$ impressions; no importance weight is truncated.

We fit a structured Brier click scorer, which gives an exact quadratic objective. Write $\mathcal D_{\rm B}$ and $\mathcal D_{\rm R}$ for the BTS and Random logs, with sizes $n_{\rm B}$ and $n_{\rm R}$. Let $Y_i\in\{0,1\}$ denote the click, let $j_i\in\{1,\ldots,36\}$ index the display-position--item-category cell, and let $e_j$ be the $j$th standard basis vector. For $w\in[0,1]^{36}$, we denote by $\widehat F_{\rm B}^{\rm IPS}$ the IPS-weighted empirical Brier risk on the BTS log and by $\widehat F_{\rm R}$ the empirical Brier risk on the independent Random-policy log:
\begin{equation}
\begin{aligned}
    \widehat F_{\rm B}^{\rm IPS}(w) &=\frac{1}{2n_{\rm B}}\sum_{i\in\mathcal D_{\rm B}} \rho_i(w_{j_i}-Y_i)^2,\\
    \widehat F_{\rm R}(w) &=\frac{1}{2n_{\rm R}}\sum_{i\in\mathcal D_{\rm R}}(w_{j_i}-Y_i)^2,\\
    g_i(w)&=\rho_i(w_{j_i}-Y_i)e_{j_i}.
\end{aligned}
\end{equation}
Thus $g_i$ is unbiased for the empirical IPS source objective $\widehat F_{\rm B}^{\rm IPS}$, while $\widehat F_{\rm R}$ is the separately collected target-policy evaluation risk. All $36$ cells have positive target frequency, so the empirical target objective is strongly convex on this feature space. We compare projected U-SGD with fixed whole-gradient clipping $h^{\rm WG}$~\eqref{eq:model-policies}.

Smaller $\lambda$ means more aggressive clipping, while $\lambda=\infty$ recovers U-SGD.  All methods return the complete post-update Polyak--Ruppert average; let $\bar w_N^{(M)}$ denote the average returned by method $M$ after $N$ sampled BTS rows.

The stepsize and clipping threshold are selected on separate source-validation trajectories before target evaluation. All methods share the selected stepsize, and the selected threshold $\lambda_{\rm val}= \OBDSelectedThreshold$ remains fixed across evaluation horizons. The Random-policy log is used only for evaluation. Appendix~\ref{app:obd-protocol} gives the complete selection protocol, including the threshold grid and the independent evaluation runs.

For any method $M$, define its target upper-tail error by
\begin{equation}\label{eq:obd-error-ratio}
\begin{aligned}
    G_M(N)&=Q_{0.95}\!\left(\widehat F_{\rm R}(\bar w_N^{(M)})- \min_{w\in[0,1]^{36}}\widehat F_{\rm R}(w)\right), \\
    R_M(N)&=\frac{G_M(N)}{G_{\rm U}(N)}.
\end{aligned}
\end{equation}
The figure reports the absolute error $G_M$, for which lower is better. Ratios $R_M$ are used only to state paired effect sizes and confidence intervals.

\begin{figure}[h]
\centering
\includegraphics[width=\textwidth]{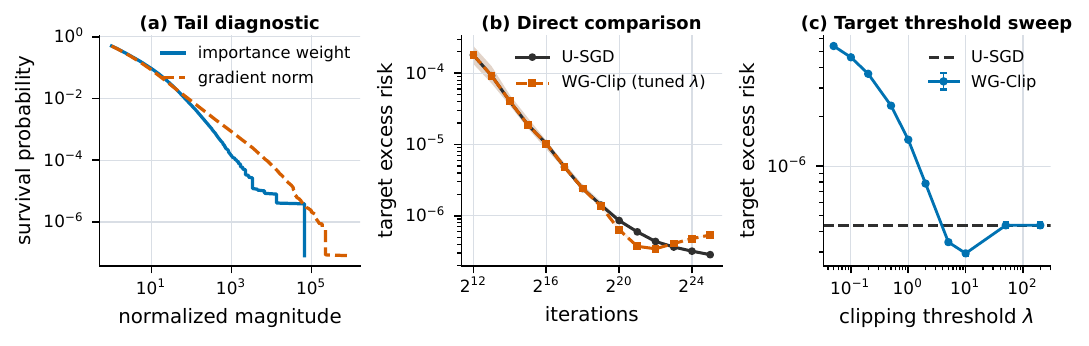}

\caption{OBD off-policy click prediction.  Lower is better in (b)--(c). (a) Empirical survival functions of the importance weight and stochastic-gradient norm at the IPS optimum, each normalized by its median; the effective-sample fraction is \OBDEffectiveSamplePercent{} and the largest weight is \OBDMaxImportanceWeight{}. (b) Direct target-error comparison: $G_M(N)$ from~\eqref{eq:obd-error-ratio} for U-SGD and WG-Clip with the single source-validation-selected threshold $\lambda_{\rm val} = \OBDSelectedThreshold$, held fixed throughout every trajectory. (c) Target-error sensitivity $G_M(\OBDFinalHorizon{})$ across every predeclared finite threshold; the dashed horizontal line is U-SGD. Points below it favor WG-Clip, whereas points above it favor U-SGD. This post-selection sweep is a diagnostic and is not used for tuning.  Bands and bars are pointwise distribution-free $95\%$ order-statistic intervals conditional on the fixed logs; paired-bootstrap intervals are reported in the text.}
\label{fig:obd-tradeoff}
\end{figure}

Figure~\ref{fig:obd-tradeoff}(b) shows the comparison as the budget grows. At the threshold-selection horizon, on the fresh evaluation trajectories, the frozen clipped method has ratio $R_{\rm Clip}=\OBDSelectedTuneRatio$ with paired $95\%$ interval $[\OBDSelectedTuneRatioLow, \OBDSelectedTuneRatioHigh]$. Thus source-only threshold selection yields a finite-horizon gain on fresh evaluation trajectories. The same fixed method is still better at $N=\OBDFinalHorizon{}$ ($R_{\rm Clip}=\OBDSelectedSweepRatio$), but the curves subsequently cross. The first displayed checkpoint after which the pointwise paired lower bound remains above one is $N=\OBDDirectCrossover$. At $N=\OBDDirectFinalHorizon{}$, the clipped-to-unclipped ratio is $\OBDDirectFinalRatio$ $[\OBDDirectFinalRatioLow,\OBDDirectFinalRatioHigh]$. Thus the method chosen without target information first helps and later loses, consistently with its shifted mean-field target becoming the dominant error.

Panel~(c) reports sensitivity to the threshold; it is evaluated separately from the selection used in panel~(b). At $N=\OBDFinalHorizon{}$, the smallest error in the displayed finite grid occurs at $\lambda=\OBDSweepBestThreshold$, with ratio $\OBDSweepBestRatio$ $[\OBDSweepBestRatioLow,\OBDSweepBestRatioHigh]$ relative to U-SGD, whereas aggressive clipping is substantially worse and very large thresholds coincide with U-SGD. Together, panels (b)--(c) show that a threshold that helps at one budget need not remain beneficial as the run grows.

\section{Conclusion}\label{sec:conclusion}

The benefit of clipping depends on the requested accuracy, confidence, and clipping bias. Under a finite second moment alone, robustification improves the confidence dependence of the ordinary untrimmed average, and bias-balanced clipping provides logarithmic confidence bounds. For $p>2$, unclipped Polyak--Ruppert averaging enters a finite-sample Gaussian-dominated regime, so clipping need not improve the leading order. Fixed whole-gradient clipping can also attenuate the signal, shift the limiting target under asymmetric noise, and change asymptotic efficiency. In the linear noise-centered setting, a growing threshold can recover the standard first-order Polyak--Ruppert limit. The bounds, exact examples, and off-policy experiment therefore support comparing clipping policies at the intended budget while accounting for both confidence gains and their effect on the target.

\bibliographystyle{unsrt}  
\bibliography{references}  

@ARTICLE{oymak_arxiv,
       author = {{Oymak}, Samet and {Soltanolkotabi}, Mahdi},
        title = "{Overparameterized Nonlinear Learning: Gradient Descent Takes the Shortest Path?}",
      journal = {arXiv e-prints},
         year = 2018,
        month = dec,
          eid = {arXiv:1812.10004},
        pages = {arXiv:1812.10004},
          doi = {10.48550/arXiv.1812.10004},
archivePrefix = {arXiv},
       eprint = {1812.10004},
 primaryClass = {cs.LG},
       adsurl = {https://ui.adsabs.harvard.edu/abs/2018arXiv181210004O}
}

@techreport{Ruppert1988,
  title={Efficient estimations from a slowly convergent Robbins-Monro process},
  author={Ruppert, David},
  year={1988},
  institution={Cornell University Operations Research and Industrial Engineering}
}

@article{PolyakJuditsky1992,
  author={Polyak, Boris T. and Juditsky, Anatoli B.},
  title={Acceleration of Stochastic Approximation by Averaging},
  journal={SIAM Journal on Control and Optimization},
  volume={30}, number={4}, pages={838--855}, year={1992}
}

@inproceedings{MouEtAl2020,
  author={Mou, Wenlong and Li, Chris Junchi and Wainwright, Martin J. and Bartlett, Peter L. and Jordan, Michael I.},
  title={On Linear Stochastic Approximation: Fine-Grained Polyak--Ruppert and Non-Asymptotic Concentration},
  booktitle={Proceedings of the 33rd Conference on Learning Theory}, series={Proceedings of Machine Learning Research},
  volume={125}, pages={2947--2997}, year={2020}
}

@article{DavisEtAl2021,
  author={Davis, Damek and Drusvyatskiy, Dmitriy and Xiao, Lin and Zhang, Junyu},
  title={From Low Probability to High Confidence in Stochastic Convex Optimization},
  journal={Journal of Machine Learning Research}, volume={22}, number={49}, pages={1--38}, year={2021}
}

@article{Catoni2012,
  author={Catoni, Olivier}, title={Challenging the Empirical Mean and Empirical Variance: A Deviation Study},
  journal={Annales de l'Institut Henri Poincar\'e, Probabilit\'es et Statistiques}, volume={48}, number={4}, pages={1148--1185}, year={2012},
  doi={10.1214/11-AIHP454}
}

@article{LugosiMendelson2019,
  author={Lugosi, Gabor and Mendelson, Shahar}, title={Sub-Gaussian Estimators of the Mean of a Random Vector},
  journal={The Annals of Statistics}, volume={47}, number={2}, pages={783--794}, year={2019}
}

@article{Nagaev2003,
  title={On probability and moment inequalities for supermartingales and martingales},
  author={Nagaev, Sergey Victorovich},
  journal={Acta Applicandae Mathematica},
  volume={79},
  number={1-2},
  pages={35--46},
  year={2003},
  publisher={Springer}
}

@article{Pinelis1994,
  author={Pinelis, Iosif}, title={Optimum Bounds for the Distributions of Martingales in Banach Spaces},
  journal={The Annals of Probability}, volume={22}, number={4}, pages={1679--1706}, year={1994}
}

@misc{MollenhauerFiedler2025,
  author={Mollenhauer, Mattes and Fiedler, Christian},
  title={Fuk--Nagaev Inequality in Smooth Banach Spaces: Optimum Bounds for Distributions of Heavy-Tailed Martingales},
  year={2025}, eprint={2512.10012}, archivePrefix={arXiv}, primaryClass={math.PR},
  note={arXiv:2512.10012}
}

@inproceedings{GorbunovDanilovaGasnikov2020,
  author={Gorbunov, Eduard and Danilova, Marina and Gasnikov, Alexander},
  title={Stochastic Optimization with Heavy-Tailed Noise via Accelerated Gradient Clipping},
  booktitle={Advances in Neural Information Processing Systems 33}, pages={15042--15053}, year={2020}
}

@inproceedings{MarshallEtAl2025,
  title={To clip or not to clip: the dynamics of SGD with gradient clipping in high-dimensions},
  author={Marshall, Noah and Xiao, Ke Liang and Agarwala, Atish and Paquette, Elliot},
  booktitle={International Conference on Learning Representations},
  volume={2025},
  pages={27381--27417},
  year={2025}
}

@inproceedings{GorbunovEtAl2024,
  author={Gorbunov, Eduard and Sadiev, Abdurakhmon and Danilova, Marina and Horv{\'a}th, Samuel and Gidel, Gauthier and Dvurechensky, Pavel and Gasnikov, Alexander and Richt{\'a}rik, Peter},
  title={High-Probability Convergence for Composite and Distributed Stochastic Minimization and Variational Inequalities with Heavy-Tailed Noise},
  booktitle={Proceedings of the 41st International Conference on Machine Learning}, series={Proceedings of Machine Learning Research}, volume={235}, pages={15951--16070}, year={2024}
}

@inproceedings{EldowaPaudice2024,
  author={Eldowa, Khaled and Paudice, Andrea}, title={General Tail Bounds for Non-Smooth Stochastic Mirror Descent},
  booktitle={Proceedings of the 27th International Conference on Artificial Intelligence and Statistics}, series={Proceedings of Machine Learning Research}, volume={238}, pages={3205--3213}, year={2024}
}

@misc{FatkhullinHueblerLan2025,
  author={Fatkhullin, Ilyas and H{\"u}bler, Florian and Lan, Guanghui},
  title={Can SGD Handle Heavy-Tailed Noise?},
  year={2025}, eprint={2508.04860}, archivePrefix={arXiv}, primaryClass={math.OC},
  note={arXiv:2508.04860}
}

@inproceedings{SadievEtAl2023,
  author={Sadiev, Abdurakhmon and Danilova, Marina and Gorbunov, Eduard and Horv{\'a}th, Samuel and Gidel, Gauthier and Dvurechensky, Pavel and Gasnikov, Alexander and Richt{\'a}rik, Peter},
  title={High-Probability Bounds for Stochastic Optimization and Variational Inequalities: The Case of Unbounded Variance},
  booktitle={Proceedings of the 40th International Conference on Machine Learning},
  series={Proceedings of Machine Learning Research}, volume={202}, pages={29563--29648}, year={2023}
}

@misc{NguyenEtAl2023,
  author={Nguyen, Ta Duy and Nguyen, Thien Hang and Ene, Alina and Nguyen, Huy Le},
  title={High Probability Convergence of Clipped-SGD under Heavy-Tailed Noise},
  year={2023}, eprint={2302.05437}, archivePrefix={arXiv}, primaryClass={math.OC},
  note={arXiv:2302.05437}
}

@misc{LiuExpectation2026,
  author={Liu, Zijian}, title={In-Expectation Convergence of Stochastic Gradient Methods under Heavy-Tailed Noise},
  year={2026}, eprint={2606.00520}, archivePrefix={arXiv}, primaryClass={cs.LG},
  note={arXiv:2606.00520}
}

@misc{HeLu2025,
  author={He, Chuan and Lu, Zhaosong}, title={Accelerated Stochastic First-Order Method for Convex Optimization under Heavy-Tailed Noise},
  year={2025}, eprint={2510.11676}, archivePrefix={arXiv}, primaryClass={math.OC},
  note={arXiv:2510.11676}
}

@inproceedings{DurmusEtAl2021,
  author={Durmus, Alain and Moulines, Eric and Naumov, Alexey and Samsonov, Sergey and Scaman, Kevin and Wai, Hoi-To},
  title={Tight High Probability Bounds for Linear Stochastic Approximation with Fixed Stepsize},
  booktitle={Advances in Neural Information Processing Systems 34},
  pages={30063--30074}, year={2021}
}

@article{DurmusEtAl2025,
  author={Durmus, Alain and Moulines, Eric and Naumov, Alexey and Samsonov, Sergey},
  title={Finite-Time High-Probability Bounds for Polyak--Ruppert Averaged Iterates of Linear Stochastic Approximation},
  journal={Mathematics of Operations Research}, volume={50}, number={2}, pages={935--964}, year={2025},
  note={Published online in 2024}, doi={10.1287/moor.2022.0179}
}

@article{samsonov2024gaussian,
  title={Gaussian approximation and multiplier bootstrap for polyak-ruppert averaged linear stochastic approximation with applications to td learning},
  author={Samsonov, Sergey and Moulines, Eric and Shao, Qi-Man and Zhang, Zhuo-Song and Naumov, Alexey},
  journal={Advances in Neural Information Processing Systems},
  volume={37},
  pages={12408--12460},
  year={2024}
}

@article{samsonov2026statistical,
  title={Statistical inference for Linear Stochastic Approximation with Markovian Noise},
  author={Samsonov, Sergey and Sheshukova, Marina and Moulines, Eric and Naumov, Alexey},
  journal={Advances in Neural Information Processing Systems},
  volume={38},
  pages={174565--174626},
  year={2026}
}

@inproceedings{SamsonovEtAl2024TD,
  author={Samsonov, Sergey and Tiapkin, Daniil and Naumov, Alexey and Moulines, Eric},
  title={Improved High-Probability Bounds for the Temporal Difference Learning Algorithm via Exponential Stability},
  booktitle={Proceedings of the 37th Conference on Learning Theory},
  series={Proceedings of Machine Learning Research}, volume={247}, pages={4511--4547}, year={2024}
}

@article{HuoChenXie2026,
  author={Huo, Dongyan Lucy and Chen, Yudong and Xie, Qiaomin},
  title={Bias and Extrapolation in Markovian Linear Stochastic Approximation with Constant Step Sizes},
  journal={Mathematics of Operations Research}, year={2026},
  note={Articles in Advance}, doi={10.1287/moor.2024.0471}
}

@book{HallHeyde1980,
  author={Hall, Peter and Heyde, Christopher C.},
  title={Martingale Limit Theory and Its Application},
  publisher={Academic Press},
  address={New York},
  year={1980}
}

@inproceedings{KoloskovaHendrikxStich2023,
  author={Koloskova, Anastasia and Hendrikx, Hadrien and Stich, Sebastian U.},
  title={Revisiting Gradient Clipping: Stochastic Bias and Tight Convergence Guarantees},
  booktitle={Proceedings of the 40th International Conference on Machine Learning},
  series={Proceedings of Machine Learning Research},
  volume={202},
  pages={17343--17363},
  year={2023},
  publisher={PMLR}
}

@inproceedings{PascanuEtAl2013,
  author={Pascanu, Razvan and Mikolov, Tomas and Bengio, Yoshua},
  title={On the Difficulty of Training Recurrent Neural Networks},
  booktitle={Proceedings of the 30th International Conference on Machine Learning},
  series={Proceedings of Machine Learning Research},
  volume={28},
  pages={1310--1318},
  year={2013}
}

@inproceedings{BrownEtAl2020,
  author={Brown, Tom B. and Mann, Benjamin and Ryder, Nick and Subbiah, Melanie and others},
  title={Language Models are Few-Shot Learners},
  booktitle={Advances in Neural Information Processing Systems},
  volume={33},
  pages={1877--1901},
  year={2020},
  url={https://papers.nips.cc/paper/2020/hash/1457c0d6bfcb4967418bfb8ac142f64a-Abstract.html}
}

@article{touvron2023llama,
  title={Llama 2: Open foundation and fine-tuned chat models},
  author={Touvron, Hugo and Martin, Louis and Stone, Kevin and Albert, Peter and Almahairi, Amjad and Babaei, Yasmine and Bashlykov, Nikolay and Batra, Soumya and Bhargava, Prajjwal and Bhosale, Shruti and others},
  journal={arXiv preprint arXiv:2307.09288},
  year={2023}
}

@article{SaitoEtAl2021OBD,
  title={Open bandit dataset and pipeline: Towards realistic and reproducible off-policy evaluation},
  author={Saito, Yuta and Aihara, Shunsuke and Matsutani, Megumi and Narita, Yusuke},
  journal={arXiv preprint arXiv:2008.07146},
  year={2020}
}

@article{Ionides2008,
  author={Ionides, Edward L.},
  title={Truncated Importance Sampling},
  journal={Journal of Computational and Graphical Statistics},
  volume={17},
  number={2},
  pages={295--311},
  year={2008},
  doi={10.1198/106186008X320456}
}

@article{BottouEtAl2013,
  author={Bottou, L\'{e}on and Peters, Jonas and Qui\~{n}onero-Candela, Joaquin and Charles, Denis X. and Chickering, D. Max and Portugaly, Elon and Ray, Dipankar and Simard, Patrice and Snelson, Ed},
  title={Counterfactual Reasoning and Learning Systems: The Example of Computational Advertising},
  journal={Journal of Machine Learning Research},
  volume={14},
  pages={3207--3260},
  year={2013},
  url={https://www.jmlr.org/papers/v14/bottou13a.html}
}
\appendix
\section*{Guide to the Appendices}

The appendices follow the main results. Each proof is kept with its statement or in the corresponding proof section.

\begin{center}
\begin{tabularx}{\linewidth}{@{}lX@{}}
\toprule
Appendix & Contents \\
\midrule
\ref{app:concentration} & Concentration tools, crossover calculations, and lower constructions. \\
\ref{app:convex-proofs} & Complete proofs for projected convex optimization. \\
\ref{app:quadratic-epochs} & Quadratic epochs, restarts, and difference-clipping certificates. \\
\ref{app:strong-proofs} & Polyak--Ruppert limits, the shifted-root example, and growing thresholds. \\
\ref{app:additional-results} & Deterministic transit and constant-step covariance. \\
\ref{app:obd-protocol} & Protocol for the OBD experiment. \\
\ref{sec:acceleration} & Hypotheses and proof targets for accelerated methods. \\
\bottomrule
\end{tabularx}
\end{center}

\section{Concentration Tools and Lower Constructions}
\label{sec:nagaev}
\label{app:concentration}
This appendix collects the concentration results, crossover calculations, and lower constructions used in the main text.  We state the results for Hilbert-valued martingale differences; scalar versions follow immediately.  Throughout, $\ell_\delta:=\log(4/\delta)$. Constants denoted by $C$ are absolute, while $C_p$ may depend only on $p$; repeated uses in distinct statements need not denote the same numerical values.

Since $\log(4m/\delta)\le(1+\log m/\log 4)\,\ell_\delta$, combining $m$ concentration events changes only numerical constants.

\subsection{A martingale Fuk--Nagaev bound}

Let $(d_k,\cF_k)_{k=1}^N$ be martingale differences in a finite-dimensional Hilbert space.  Define predictable moment budgets
\begin{equation}
    V_N = \sum_{k=1}^N \E\left[\norm{d_k}^2 \mid\cF_{k-1}\right], \quad A_{p,N} = \sum_{k=1}^N\E\left[\norm{d_k}^p \mid\cF_{k-1}\right].
\end{equation}

\begin{theorem}[martingale Fuk--Nagaev inequality]
\label{thm:fuk-nagaev}
Let $p>2$ and suppose $V_N\le v^2$ and $A_{p,N}\le a_p^p$ almost surely.  Then, for every $\delta\in(0,1)$,
\begin{equation}
    \norm{\sum_{k=1}^N d_k} \le C \, v\sqrt{\ell_\delta} +C_p a_p\delta^{-1/p}
\end{equation}
with probability at least $1-\delta$, where $\ell_\delta=\log(4/\delta)$.  A corresponding tail form states that, for every $t>0$,
\begin{equation}
    \Pp \left(\max_{j\le N} \norm{\sum_{k=1}^j d_k} > t \right) \le 2 \exp \left(-\frac{t^2}{C v^2}\right) +\frac{C_p^p a_p^p}{t^p}.
\end{equation}
\end{theorem}

The structure, rather than a particular numerical constant, is essential: the exponential term describes the accumulation of many moderate increments, while the power term is of the same order as the probability that one increment alone exceeds $t$. Results of this form go back to S.~V.~Nagaev and martingale extensions~\cite{Nagaev2003}; sharp Banach-space formulations are based on Pinelis' smooth-norm inequalities~\cite{Pinelis1994}. 
\begin{proof}
Theorem~2.2 and Corollary~2.3 of Mollenhauer and Fiedler~\cite{MollenhauerFiedler2025} apply under essential-supremum conditional moment budgets in a $(2,D)$-smooth Banach space.  A Hilbert space is $(2,1)$-smooth, and their notation maps to ours as
\[
    q=p,\quad D=1,\quad \sigma^2=V_N\le v^2,\quad C_q^q=A_{p,N}\le a_p^p,\quad M_j=\sum_{k\le j}d_k.
\]
Their confidence coefficient is
\[
    c_{p,1}=\frac{1}{2p}+\min\{1/p,1/5\}+1 +\mathbf1_{\{p>3\}}\frac p3.
\]
The constants are therefore dimension-free in a Hilbert space, remain bounded as $p\downarrow2$, and grow at most linearly in $p$ for large $p$. The symbols $C,C_p$ absorb the source factors $\sqrt2$, $2^{1/p}$, and $c_{p,1}$.
\end{proof}

At the endpoint $p=2$, the decomposition does not provide two different scales.  Chebyshev's inequality gives
\begin{equation}
    \norm{\sum_{k=1}^N d_k}\le v\delta^{-1/2}
\end{equation}
with probability at least $1-\delta$. Without a stronger assumption or a robust transformation, the factor $\delta^{-1/2}$ cannot be replaced uniformly by $\sqrt{\log(1/\delta)}$ for a single empirical average; see Catoni~\cite[Proposition~6.2]{Catoni2012} for an explicit fixed-variance lower bound.

\begin{corollary}[weighted stochastic-gradient noise]
\label{cor:weighted-nagaev}
Under Assumption~\ref{ass:noise}, let $w_k$ be predictable scalar weights.  For $p>2$, with probability at least $1-\delta$,
\begin{equation}\label{eq:weighted-nagaev}
    \norm{\sum_{k=1}^N w_k \zeta_k} \le C\sigma_2 \sqrt{\ell_\delta\sum_{k=1}^Nw_k^2} + C_p \sigma_p \delta^{-1/p} \left(\sum_{k=1}^N \abs{w_k}^p\right)^{1/p}.
\end{equation}
In particular, for $w_k=1/N$,
\begin{equation}\label{eq:unclipped-mean-radius}
    \norm{\frac1N\sum_{k=1}^N\zeta_k} \le \; r_{\rm U}^{\rm mean}(N,\delta) := C \sigma_2 \sqrt{\frac{\ell_\delta}{N}} + C_p\sigma_p \delta^{-1/p}N^{-(1-(1/p))}
\end{equation}
with probability at least $1-\delta$.
\end{corollary}

\begin{proof}
Take $d_k=w_k\zeta_k$ in Theorem~\ref{thm:fuk-nagaev}. Predictability of $w_k$ and Assumption~\ref{ass:noise} give
\[
    V_N\le \sigma_2^2\sum_{k=1}^Nw_k^2, \quad A_{p,N}\le \sigma_p^p\sum_{k=1}^N|w_k|^p.
\]
Substitution proves \eqref{eq:weighted-nagaev}.  For $w_k=1/N$, the two weight norms are $N^{-1/2}$ and $N^{-(1-1/p)}$, which gives \eqref{eq:unclipped-mean-radius}.
\end{proof}

For fixed $\delta$, the second term in \eqref{eq:unclipped-mean-radius} is lower order in $N$ exactly when $p>2$. Proposition~\ref{prop:one-outlier-lower} shows that its $(\sigma_p,N,\delta)$ dependence is sharp for the empirical-mean mechanism, but not uniformly over classes with both $\sigma_2$ and $\sigma_p$ fixed.

\subsection{Bias and concentration after clipping}
\label{app:clipping-bias-variance}

The next elementary lemma exposes all three effects of clipping.

\begin{lemma}[clipped conditional mean]
\label{lem:clipped-mean}
Let $Y$ be a random vector with conditional mean $m=\E[Y\mid\cG]$.

\begin{enumerate}[label=\textnormal{(\roman*)}]
\item If $\E[\norm{Y}^p\mid\cG]\le M_p^p$, then
\begin{equation}\label{eq:clip-bias-raw}
    \norm{\E[\clip_\lambda(Y)\mid\cG]-m} \le \frac{M_p^p}{\lambda^{p-1}}.
\end{equation}
\item If $Y=m+Z$, $\E[Z\mid\cG]=0$, and one clips $Z$ rather than $Y$, then
\begin{equation}\label{eq:clip-bias-centered}
    \norm{\E[\clip_\lambda(Z)\mid\cG]} \le \frac{\E[\norm{Z}^p\mid\cG]}{\lambda^{p-1}}.
\end{equation}
Thus the raw scale $M_p$ is replaced by the central scale $\sigma_p$.
\item The projection $\clip_\lambda (\cdot)$ is nonexpansive. Consequently, for an independent conditional copy $Y'$,
\begin{equation}\label{eq:variance-contraction}
    \E\!\left[\norm{\clip_\lambda(Y)-\E[\clip_\lambda(Y)\mid\cG]}^2\mid\cG\right] \le \frac12 \E\left[\norm{Y-Y'}^2 \mid \cG \right].
\end{equation}
For $Y=m+Z$ this is at most $\E\left[\norm{Z}^2 \mid \cG \right]$.
\end{enumerate}
\end{lemma}

\begin{proof}
For every vector $y$,
\begin{equation}
    \norm{\clip_\lambda(y)-y} = (\norm y-\lambda)_+ \le \norm y\mathbf1\{\norm y>\lambda\} \le \frac{\norm y^p}{\lambda^{p-1}}.
\end{equation}
Conditional Jensen's inequality yields \eqref{eq:clip-bias-raw}; the same argument applied to $Z$ proves \eqref{eq:clip-bias-centered}.

For any square-integrable random vector $U$ and an independent conditional copy $U'$,
\begin{equation}
    \E[\norm{U-\E[U\mid\cG]}^2\mid\cG] =\frac12\E[\norm{U-U'}^2\mid\cG]. \label{eq:variance-copy}
\end{equation}
Indeed, conditional independence and equality of the conditional laws give
\[
    \E[\norm{U-U'}^2\mid\cG] =2\E[\norm U^2\mid\cG]-2\norm{\E[U\mid\cG]}^2.
\]
Take $U=\clip_\lambda(Y)$.  Since Euclidean projection is nonexpansive,
\[
    \norm{\clip_\lambda(Y)-\clip_\lambda(Y')}\le\norm{Y-Y'},
\]
which proves \eqref{eq:variance-contraction}.  If $Y=m+Z$, the right-hand side of \eqref{eq:variance-copy} is $\E[\norm Z^2\mid\cG]$.
\end{proof}

Combining these bias and variance bounds with bounded-increment martingale concentration yields the following control of clipped averages.

\begin{proposition}[concentration of clipped averages]
\label{prop:clipped-average}
Let $m_k=\E[Y_k\mid\cF_{k-1}]$, and suppose that, almost surely for every $k$, \(\E[\norm{Y_k-m_k}^2\mid\cF_{k-1}]\le\sigma_2^2\) and \(\E[\norm{Y_k}^p\mid\cF_{k-1}]\le M_p^p\).  Then, for every $\lambda>0$ and $\delta\in(0,1)$,
\begin{equation}\label{eq:clip-mean-general}
    \norm{\frac1N\sum_{k=1}^N\bigl(\clip_\lambda(Y_k)-m_k\bigr)} \le r_{\rm C}^{\rm mean}(N,\delta;\lambda) :=C\sigma_2\sqrt{\frac{\ell_\delta}{N}} +C\lambda\frac{\ell_\delta}{N} +\frac{M_p^p}{\lambda^{p-1}}
\end{equation}
with probability at least $1-\delta$.  The non-Gaussian terms are balanced, up to constants, by
\begin{equation}\label{eq:clip-optimal-mean}
    \lambda_\star\asymp M_p\left(\frac{N}{\ell_\delta}\right)^{1/p}, \quad r_{\rm C}^{\rm mean}(N,\delta) \lesssim \sigma_2\sqrt{\frac{\ell_\delta}{N}} + M_p\left(\frac{\ell_\delta}{N}\right)^{1-1/p}.
\end{equation}
\end{proposition}

\phantomsection
\label{app:clipped-freedman-proof}
\begin{proof}
Let
\begin{equation}
    u_k=\clip_\lambda(Y_k) -\E[\clip_\lambda(Y_k)\mid\cF_{k-1}].
\end{equation}
Then $(u_k)$ is a martingale-difference sequence and $\norm{u_k}\le2\lambda$.  Lemma~\ref{lem:clipped-mean}(iii) gives
\[
    \E[\norm{u_k}^2\mid\cF_{k-1}] \le \E[\norm{Y_k-m_k}^2\mid\cF_{k-1}] \le \sigma_2^2.
\]
Apply Pinelis' Bennett--Freedman inequality for martingales in $2$-smooth Banach spaces~\cite[Theorem~3.4]{Pinelis1994}, with increment bound $2\lambda$ and quadratic-variation bound $N\sigma_2^2$, to obtain
\begin{equation}
    \norm{\sum_{k=1}^Nu_k} \le C\sigma_2\sqrt{N\ell_\delta}+C\lambda\ell_\delta 
    \label{eq:clipped-freedman-sum}
\end{equation}
with probability at least $1-\delta$.  Divide by $N$ and use Lemma~\ref{lem:clipped-mean}(i) for the average conditional bias to obtain \eqref{eq:clip-mean-general}.

To optimize the non-Gaussian part, minimize
\begin{equation}
    \phi(\lambda)=c_1\lambda\frac{\ell_\delta}{N} +c_2\frac{M_p^p}{\lambda^{p-1}}.
\end{equation}
The stationary equation is
\[
    c_1\frac{\ell_\delta}{N} =(p-1)c_2M_p^p\lambda^{-p},
\]
and hence $\lambda\asymp M_p(N/\ell_\delta)^{1/p}$ and $\phi(\lambda)\asymp M_p(\ell_\delta/N)^{1-1/p}$, proving \eqref{eq:clip-optimal-mean}.
\end{proof}

For noise-centered clipping $M_p=\sigma_p$; for whole-gradient clipping the valid uniform choice is $M_p=G_p$.

\subsection{Crossover boundaries and their interpretation}
\label{app:crossover-algebra}

There are two useful comparisons.  First, compare the one-big-jump term of U-SGD with the Gaussian core.

\begin{proposition}[Gaussian-dominance boundary]
\label{prop:gaussian-boundary}
Let $p>2$, $\chi_p=\sigma_p/\sigma_2$,  and define $K_p:=\left(\frac{C_p}{C}\right)^p$, where $C$ and $C_p$ are the two coefficients displayed in \eqref{eq:unclipped-mean-radius}. The polynomial term in \eqref{eq:unclipped-mean-radius} is no larger than the Gaussian term whenever
\begin{equation}\label{eq:gaussian-boundary}
  \delta \, \ell_\delta^{p/2} \ge K_p \chi_p^p N^{-(p-2)/2}.
\end{equation}

\end{proposition}

\begin{proof}
The one-big-jump term is no larger than the Gaussian term if
\[
    C_p\sigma_p\delta^{-1/p}N^{-(1-1/p)} \le C\sigma_2\ell_\delta^{1/2}N^{-1/2}.
\]
Divide by $\sigma_2N^{-1/2}$ and raise both sides to the $p$th power. This gives \eqref{eq:gaussian-boundary} with $K_p=(C_p/C)^p$.
\end{proof}

For fixed $\delta$ and $\chi_p$, the Gaussian-dominance condition \eqref{eq:gaussian-boundary} eventually holds for every $p>2$, whereas at $p=2$ increasing $N$ does not improve the comparison.

For completeness, the same algebra proves the accuracy-normalized conditions in Proposition~\ref{prop:accuracy-boundary}.  Substituting $N\asymp\rho^{-2}\ell_\delta$ into \eqref{eq:gaussian-boundary} gives \eqref{eq:accuracy-U-boundary}. Balancing the centered-clipping correction with the Gaussian term gives $\chi_p^p\rho^{p-2}\lesssim1$, which is \eqref{eq:accuracy-C-boundary}.

Second, compare only the non-Gaussian corrections of the unclipped and centered-clipped averages. Ignoring universal constants,
\begin{equation}
    \frac{\sigma_p\delta^{-1/p}N^{-(1-1/p)}} {\sigma_p(\ell_\delta/N)^{1-1/p}} = \left(\frac{1}{\delta\ell_\delta^{p-1}}\right)^{1/p}.
\end{equation}
Thus finite-threshold clipping improves this correction in the very-high-confidence region $\delta\ell_\delta^{p-1}<1$, while the unclipped Nagaev correction can be smaller at moderate confidence and sufficiently large $p$. This comparison is independent of $N$ because both corrections have the same homogeneity after the clipping-bias upper bound has been balanced.  Their importance relative to the Gaussian core, however, decays with $N$ when $p>2$.

\begin{center}
\begin{minipage}{.98\textwidth}
\captionsetup{hypcap=false}
\captionof{table}{
Finite-sample guidance after the deterministic optimization error has been reduced below the target accuracy. The conclusions concern the guarantees established in this paper; instance-wise constants can still depend on the noise distribution.
}
\label{tab:decision}
\small
\setlength{\tabcolsep}{4pt}
\begin{tabularx}{\textwidth}{
    >{\raggedright\arraybackslash}p{.23\textwidth}
    >{\raggedright\arraybackslash}p{.25\textwidth}
    X}
\toprule
Regime & Conclusion & Reason \\
\midrule

Only $p=2$ is known, and a distribution-free $1-\delta$ guarantee is required & Use a target-preserving robustification & The ordinary untrimmed average has polynomial confidence dependence. Clipping is one option, but robust aggregation or robust mean estimation can provide the same type of confidence improvement.
\\
$p>2$ and $\chi_p^p\rho^{p-2} \gtrsim \delta\ell_\delta^{p-1}$ & Finite clipping can improve the confidence correction if its bias is controlled & The heavy-tail correction of U-SGD is not yet guaranteed to be lower order. For whole-gradient clipping, the direct comparison must also account for $G_p/\sigma_p$.
\\
$p>2$ and $\chi_p^p\rho^{p-2} \lesssim \delta\ell_\delta^{p-1}$ & U-SGD is already Gaussian dominated & Finite clipping does not improve the leading order, although it may change an instance-specific constant.
\\
Large signal-to-noise ratio or asymmetric noise & Keep U-SGD as an explicit candidate & Whole-gradient clipping can truncate the optimization signal or change the limiting target. A finite threshold should therefore be used only after checking signal attenuation and clipping bias.
\\
\bottomrule
\end{tabularx}
\end{minipage}
\end{center}

The table describes the distribution-free comparison. Fixed-law behavior can still differ: a finite threshold can improve a symmetric heavy-tailed location problem, whereas U-SGD is asymptotically efficient in the scalar Gaussian case, and fixed whole-gradient clipping can shift the limiting point under asymmetric noise.

\subsection{The endpoint \texorpdfstring{$p=2$}{p=2} and matching lower constructions}

A simple scalar construction makes the obstruction transparent.  For each $N$ and $\delta$, choose a centered distribution that equals a value of order $\sigma_2\sqrt{N/\delta}$ with probability of order $\delta/N$, and uses a small compensating value otherwise.  Its variance is of order $\sigma_2^2$.  With probability of order $\delta$, at least one of the $N$ observations takes this exceptional value; conditional on this event, the empirical mean has magnitude of order $\sigma_2/\sqrt{N\delta}$.  Therefore no distribution-free confidence radius of order $\sigma_2\sqrt{\log(1/\delta)/N}$ can hold for the ordinary mean under only a second-moment bound.  Corollary~\ref{cor:quadratic-usgd-all-p-lower} embeds this mean exactly into a one-dimensional quadratic U-SGD recursion with Polyak--Ruppert averaging.  If logarithmic confidence is required uniformly over the finite-variance class, some additional robustification (clipping, median-of-means, Catoni estimation, or confidence boosting) is needed.

For $p>2$, the same construction is constrained by the higher moment: the exceptional value cannot simultaneously be as large and as frequent.  Its contribution becomes $N^{-(1-(1/p))}\delta^{-1/p}$, matching the second term of \eqref{eq:unclipped-mean-radius} for this mechanism.  This is the probabilistic reason why U-SGD re-enters the comparison for $p>2$.

The following construction formalizes this argument for every $p\ge2$.

\begin{proposition}[one-outlier lower bound for a finite $p$-th moment]
\label{prop:one-outlier-lower}
Fix $p\ge2$, $N\ge2$, and $0<\delta\le1/2$.  There is a centered scalar law
with $\E\abs X^p=\sigma_p^p$ such that, for i.i.d. copies,
\begin{equation}
    \Pp\left(\abs{\frac1N\sum_{i=1}^NX_i}\ge c_p\sigma_p\delta^{-1/p}N^{-(1-1/p)}\right)\ge c\delta.
    \label{eq:general-one-outlier-lower}
\end{equation}
For $p=2$, the same law has variance $\sigma_2^2$ and the radius becomes $c\sigma_2/\sqrt{N\delta}$.
\end{proposition}

\begin{proof}
Set $q=\delta/(4N)$ and let
\begin{equation}
    X=\begin{cases}
    A,&\text{with probability }q,\\
    -qA/(1-q),&\text{with probability }1-q,
    \end{cases}
    \qquad
    A=\sigma_p\left(q+\frac{q^p}{(1-q)^{p-1}}\right)^{-1/p}.
    \label{eq:general-p-lower-law}
\end{equation}
Then $\E X=0$ and $\E\abs X^p=\sigma_p^p$.  Let $E$ be the event that exactly one of $X_1,\ldots,X_N$ equals $A$.  Since $Nq\le1/8$,
\begin{equation}
    \Pp(E)=Nq(1-q)^{N-1}\ge c\delta.
\end{equation}
On $E$,
\begin{equation}
    \frac1N\sum_{i=1}^NX_i=\frac{A(1-Nq)}{N(1-q)}.
\end{equation}
For $q\le1/8$, the factor defining $A$ is comparable to $q$, while $(1-Nq)/(1-q)$ is bounded below.  The right-hand side is therefore at least $c_p\sigma_pq^{-1/p}/N$, which is the scale in \eqref{eq:general-one-outlier-lower}.
\end{proof}

\begin{corollary}[an exact quadratic U-SGD lower bound for every $p\ge2$]
\label{cor:quadratic-usgd-all-p-lower}
Consider the one-dimensional quadratic
\begin{equation}
    F(x)=\frac{x^2}{2}, \qquad g(x,X)=x+X,
\end{equation}
run U-SGD with the admissible constant step $\eta_k\equiv1$ and output $\bar x_N=N^{-1}\sum_{k=1}^Nx_k$. For the law in \eqref{eq:general-p-lower-law}, for every $p\ge2$,
\begin{equation}
    \Pp\left(F(\bar x_N)-F(0) \ge c_p\sigma_p^2\delta^{-2/p}N^{-2(1-1/p)}\right) \ge c'\delta.
    \label{eq:all-p-quadratic-gap-lower}
\end{equation}
In particular, at $p=2$,
\begin{equation}
    \Pp\left(F(\bar x_N)-F(0) \ge c\frac{\sigma_2^2}{N\delta}\right)\ge c'\delta.
    \label{eq:p2-quadratic-gap-lower}
\end{equation}
Thus the one-jump objective term is sharp for this explicit untrimmed single-run Polyak--Ruppert mechanism. The statement does not cover median-of-means, Catoni-type final aggregation, or confidence boosting across independent runs.
\end{corollary}

\begin{proof}
Index the update from $x_{k-1}$ to $x_k$.  The recursion is exact: $x_k=x_{k-1}-(x_{k-1}+X_k)=-X_k$. Hence $\bar x_N=-N^{-1}\sum_{k=1}^NX_k$.  Proposition~\ref{prop:one-outlier-lower}, followed by $F(x)=x^2/2$, proves \eqref{eq:all-p-quadratic-gap-lower}; \eqref{eq:p2-quadratic-gap-lower} is its $p=2$ specialization.
\end{proof}

\section{Proofs for Convex Optimization}
\label{app:convex-proofs}

Throughout this appendix, write $\E_{k-1}[\cdot]:=\E[\cdot\mid\cF_{k-1}]$ and use $\ell_\delta=\log(12/\delta)$. Constants $C$ are absolute, whereas $C_p$ may depend only on $p$. All stepsizes and clipping thresholds considered below are deterministic.

\subsection{The common projected inequality}
\label{app:projected-inequality}

We first derive the common projected inequality. For any direction $h_k$, write
\begin{equation}\label{eq:generic-decomposition}
    h_k=s_k+b_k+u_k,\quad \E_{k-1}[u_k]=0,
\end{equation}
where $b_k$ is a predictable bias. Nonexpansiveness of projection gives
\begin{align}
    \norm{x_{k+1}-x^\star}^2
    &\le \norm{x_k-x^\star-\eta h_k}^2
    \notag\\
    &=\norm{x_k-x^\star}^2 -2\eta\ip{s_k}{x_k-x^\star} -2\eta\ip{b_k+u_k}{x_k-x^\star} +\eta^2\norm{h_k}^2.
    \label{eq:generic-projection}
\end{align}
Since $F(x_k)-F(x^\star)\le\ip{s_k}{x_k-x^\star}$, we obtain
\begin{align}
    F(x_k)-F(x^\star) \le{}& \frac{\norm{x_k-x^\star}^2 -\norm{x_{k+1}-x^\star}^2}{2\eta} +\frac{\eta}{2}\norm{h_k}^2
    \notag\\
    &-\ip{u_k}{x_k-x^\star} -\ip{b_k}{x_k-x^\star}.
    \label{eq:generic-convex-one-step}
\end{align}
Summing, dividing by $N$, and applying Jensen's inequality proves \eqref{eq:convex-master}. Bounding the bias term using $\norm{x_k-x^\star}\le D$ also gives
\begin{align}
    F(\bar x_N)-F(x^\star) \le{}& \frac{R_0^2}{2\eta N} +\frac{\eta}{2N}\sum_{k=1}^N\norm{h_k}^2 \notag\\
    &-\frac1N\sum_{k=1}^N\ip{u_k}{x_k-x^\star} +\frac{D}{N}\sum_{k=1}^N\norm{b_k}.
    \label{eq:generic-convex-sum}
\end{align}

For U-SGD, $h_k=g_k$, $b_k=0$, and $u_k=\zeta_k$. Define
\begin{equation}\label{eq:convex-scalar-md}
    d_k:=\ip{\zeta_k}{x_k-x^\star}.
\end{equation}
Predictability of $x_k$ gives
\[
    \E_{k-1}[d_k]=0, \quad \E_{k-1}[d_k^2]\le D^2\sigma_2^2, \quad \E_{k-1}[|d_k|^p]\le D^p\sigma_p^p.
\]
For $p>2$, Theorem~\ref{thm:fuk-nagaev}, with failure probability $\delta/3$, therefore yields
\begin{equation}\label{eq:convex-md-bound}
    \frac1N\left|\sum_{k=1}^N d_k\right| \le CD\sigma_2\sqrt{\frac{\ell_\delta}{N}} +C_pD\sigma_p\delta^{-1/p}N^{-(1-1/p)}.
\end{equation}
The deterministic moment budgets used here are $ND^2\sigma_2^2$ and $ND^p\sigma_p^p$.

\subsection{Proof of Theorem~\ref{thm:convex-unclipped}}
\label{proof_U-SGD}

\begin{proof}
We first separate the predictable signal from the noise in the squared-gradient term. Since $\E_{k-1}[\zeta_k]=0$,
\begin{equation}\label{eq:convex-signal-bound-app}
    \norm{s_k}^2+\E_{k-1}\norm{\zeta_k}^2 =\E_{k-1}\norm{g_k}^2 \le G_2^2.
\end{equation}
In particular, $\norm{s_k}\le G_2$.

Define $Z_k:=\norm{\zeta_k}^2-\E_{k-1}\norm{\zeta_k}^2$
and $c_k:=\ip{s_k}{\zeta_k}$.
Both sequences are scalar martingale differences, and
\begin{equation}\label{eq:convex-energy-decomposition-app}
    \norm{g_k}^2 = \E_{k-1}\norm{g_k}^2+2c_k+Z_k.
\end{equation}

For the cross term, predictability of $s_k$ and
\eqref{eq:convex-signal-bound-app} imply
\[
    \E_{k-1}[c_k^2]\le G_2^2\sigma_2^2, \quad \E_{k-1}[|c_k|^p]\le G_2^p\sigma_p^p.
\]
Applying Theorem~\ref{thm:fuk-nagaev} with failure probability $\delta/3$ gives
\begin{equation}\label{eq:convex-cross-bound}
    \frac1N\left|\sum_{k=1}^N c_k\right| \le CG_2\sigma_2\sqrt{\frac{\ell_\delta}{N}} +C_pG_2\sigma_p\delta^{-1/p}N^{-(1-1/p)}.
\end{equation}

We next control the centered squared noise. Put $q=p/2>1$. Conditional Jensen's inequality gives
\begin{align}
    \E_{k-1}[|Z_k|^q] &\le
    2^{q-1}\left(\E_{k-1}\norm{\zeta_k}^{2q} +\bigl(\E_{k-1}\norm{\zeta_k}^2\bigr)^q \right)
    \notag\\
    &\le 2^q\E_{k-1}\norm{\zeta_k}^{2q} \le C_p\sigma_p^p.
    \label{eq:energy-q-moment}
\end{align}

If $2<p\le4$, then $1<q\le2$. The martingale von Bahr--Esseen inequality yields
\begin{equation}\label{eq:vbe-energy}
    \E\left|\sum_{k=1}^N Z_k\right|^q \le C_q\sum_{k=1}^N\E|Z_k|^q \le C_pN\sigma_p^p.
\end{equation}
For completeness, the first inequality follows by applying
\[
    |a+b|^q \le |a|^q +q|a|^{q-1}\operatorname{sgn}(a)b +C_q|b|^q
\]
to the partial sums and taking conditional expectations. The linear term vanishes because $Z_k$ is a martingale difference.

Markov's inequality applied to \eqref{eq:vbe-energy} gives, with probability at least $1-\delta/3$,
\begin{equation}\label{eq:energy-low-p}
    \frac1N\left|\sum_{k=1}^N Z_k\right| \le C_p\sigma_p^2\delta^{-2/p}N^{-(1-2/p)}, \quad 2<p\le4.
\end{equation}

If $p>4$, then $q=p/2>2$. Moreover,
\[
    \E_{k-1}[Z_k^2] \le \E_{k-1}\norm{\zeta_k}^4 \le \bigl(\E_{k-1}\norm{\zeta_k}^p\bigr)^{4/p} \le \sigma_p^4.
\]
Thus Theorem~\ref{thm:fuk-nagaev}, applied at moment order $q$ with deterministic budgets $N\sigma_p^4$ and $C_pN\sigma_p^p$, gives, with probability at least $1-\delta/3$,
\begin{equation}\label{eq:energy-high-p}
    \frac1N\left|\sum_{k=1}^N Z_k\right| \le C\sigma_p^2\sqrt{\frac{\ell_\delta}{N}} +C_p\sigma_p^2\delta^{-2/p}N^{-(1-2/p)}.
\end{equation}
Consequently, in either case,
\[
    \frac1N\left|\sum_{k=1}^N Z_k\right| \le \mathcal Q_{p,N}(\delta)
\]
with probability at least $1-\delta/3$.

Combining \eqref{eq:convex-energy-decomposition-app}, \eqref{eq:convex-cross-bound}, and the bound on $Z_k$ yields
\begin{align}
    \frac1N\sum_{k=1}^N\norm{g_k}^2 \le{}&
    G_2^2 +CG_2\sigma_2\sqrt{\frac{\ell_\delta}{N}}
    \notag\\
    &+C_pG_2\sigma_p\delta^{-1/p}N^{-(1-1/p)} +\mathcal Q_{p,N}(\delta).
    \label{eq:energy-final}
\end{align}
Intersect this event with the event in \eqref{eq:convex-md-bound}. The three concentration events have total failure probability at most $\delta$. Substitution into \eqref{eq:generic-convex-sum} gives the arbitrary-stepsize bound
\begin{align}
    F(\bar x_N)-F(x^\star) \le{}&
    \frac{R_0^2}{2\eta N} +\frac{\eta G_2^2}{2} +C(D+\eta G_2)\sigma_2 \sqrt{\frac{\ell_\delta}{N}} \notag\\
    &+C_p(D+\eta G_2)\sigma_p \delta^{-1/p}N^{-(1-1/p)} +\frac{\eta}{2}\mathcal Q_{p,N}(\delta).
    \label{eq:convex-unclipped-general}
\end{align}

Finally, choose $\eta=R_0/(G_2\sqrt N)$. Then
\[
    \frac{R_0^2}{2\eta N}+\frac{\eta G_2^2}{2} =\frac{R_0G_2}{\sqrt N}, \quad D+\eta G_2 =D+\frac{R_0}{\sqrt N} \le2D.
\]
Substituting these relations into \eqref{eq:convex-unclipped-general} proves \eqref{eq:convex-unclipped-tuned}.
\end{proof}

\subsection{Proof of Theorem~\ref{thm:convex-clipped}}
\label{proof_WholeGradientClipping}

\begin{proof}
We first derive an arbitrary-parameter bound. This part of the argument is valid for every $p\ge2$.

Set $h_k=\clip_\lambda(g_k)$ and define
\begin{equation}\label{eq:clip-decomposition}
    b_k(\lambda):=\E_{k-1}[h_k]-s_k, \quad \widetilde\zeta_k(\lambda) :=h_k-\E_{k-1}[h_k].
\end{equation}
Thus \eqref{eq:generic-decomposition} holds with $b_k=b_k(\lambda)$ and $u_k=\widetilde\zeta_k(\lambda)$.

Lemma~\ref{lem:clipped-mean} gives
\begin{equation}\label{eq:clip-bias-variance}
    \norm{b_k(\lambda)} \le\frac{G_p^p}{\lambda^{p-1}}, \quad \E_{k-1}\norm{\widetilde\zeta_k(\lambda)}^2 \le\sigma_2^2.
\end{equation}
Since $\norm{h_k}\le\lambda$, we also have
\begin{equation}\label{eq:clipped-centered-increment-bound}
    \norm{\widetilde\zeta_k(\lambda)} \le2\lambda.
\end{equation}
The contribution of the predictable bias is therefore bounded by
\begin{equation}\label{eq:convex-clip-bias-proof}
    \frac{D}{N}\sum_{k=1}^N\norm{b_k(\lambda)} \le\frac{DG_p^p}{\lambda^{p-1}}.
\end{equation}

Define the scalar martingale differences $\widetilde d_k :=\ip{\widetilde\zeta_k(\lambda)}{x_k-x^\star}$.
Their increments and conditional variances satisfy
\[
    |\widetilde d_k|\le2D\lambda, \quad \E_{k-1}[\widetilde d_k^2]\le D^2\sigma_2^2.
\]
The Bernstein--Freedman argument used for \eqref{eq:clipped-freedman-sum}, applied to these scalar increments, gives, with probability at least $1-\delta/3$,
\begin{equation}\label{eq:clip-scalar-freedman}
    \frac1N\left|\sum_{k=1}^N\widetilde d_k\right| \le CD\sigma_2\sqrt{\frac{\ell_\delta}{N}} +CD\lambda\frac{\ell_\delta}{N}.
\end{equation}

For the squared-gradient term, let
\[
    W_k:=\norm{h_k}^2-\E_{k-1}\norm{h_k}^2.
\]
Since $0\le\norm{h_k}^2\le\lambda^2$, we have $|W_k|\le\lambda^2$. Moreover,
\begin{align}
    \E_{k-1}[W_k^2] &\le\E_{k-1}\norm{h_k}^4 \notag\\
    &\le\lambda^2\E_{k-1}\norm{h_k}^2 \le\lambda^2G_2^2.
    \label{eq:clip-energy-variance}
\end{align}
Here we used $\norm{h_k}\le\norm{g_k}$. Bernstein--Freedman concentration, together with $\E_{k-1}\norm{h_k}^2\le G_2^2$, gives, with probability at least $1-\delta/3$,
\begin{equation}\label{eq:clip-energy-final}
    \frac1N\sum_{k=1}^N\norm{h_k}^2 \le G_2^2 +CG_2\lambda\sqrt{\frac{\ell_\delta}{N}} +C\lambda^2\frac{\ell_\delta}{N}.
\end{equation}

The intersection of the events in \eqref{eq:clip-scalar-freedman} and \eqref{eq:clip-energy-final} has probability at least $1-2\delta/3$, hence at least $1-\delta$. Combining these inequalities with \eqref{eq:generic-convex-sum} and \eqref{eq:convex-clip-bias-proof} yields
\begin{align}
    F(\bar x_N)-F(x^\star) \le{}& \frac{R_0^2}{2\eta N} +\frac{\eta G_2^2}{2} +CD\sigma_2\sqrt{\frac{\ell_\delta}{N}} \notag\\
    &+CD\lambda\frac{\ell_\delta}{N} +\frac{DG_p^p}{\lambda^{p-1}} \notag\\
    &+C\eta G_2\lambda\sqrt{\frac{\ell_\delta}{N}} +C\eta\lambda^2\frac{\ell_\delta}{N}.
    \label{eq:convex-clipped-general}
\end{align}

Now let $p>2$ and choose
\[
    \eta=\frac{R_0}{G_2\sqrt N}, \quad \lambda=c_pG_p(N/\ell_\delta)^{1/p},
\]
where $c_p>0$ depends only on $p$. The first two terms in \eqref{eq:convex-clipped-general} sum to $R_0G_2/\sqrt N$. Also,
\begin{align}
    D\lambda\frac{\ell_\delta}{N} +\frac{DG_p^p}{\lambda^{p-1}} &\le C_pDG_p\ell_\delta^{1-1/p}N^{-(1-1/p)},\\ 
    \eta G_2\lambda\sqrt{\frac{\ell_\delta}{N}}
    &= c_pR_0G_p\ell_\delta^{1/2-1/p}N^{-(1-1/p)},\\
    \eta\lambda^2\frac{\ell_\delta}{N}
    &= c_p^2\frac{R_0G_p^2}{G_2}
    \ell_\delta^{1-2/p}N^{-(3/2-2/p)}.
\end{align}
Because $R_0\le D$ and $\ell_\delta\ge1$, the middle expression is at most
\[
    c_pDG_p\ell_\delta^{1-1/p}N^{-(1-1/p)}.
\]
Substitution into \eqref{eq:convex-clipped-general} proves \eqref{eq:convex-clipped-tuned}.
\end{proof}

\subsection{Proof of \eqref{eq:convex-p2-optimal}}
\label{proof_convex-p2-optimal}

\begin{proof}
The scalar martingale differences $d_k$ defined in \eqref{eq:convex-scalar-md} satisfy
\[
    \E\left|\sum_{k=1}^N d_k\right|^2 =\sum_{k=1}^N\E[d_k^2] \le ND^2\sigma_2^2.
\]
Chebyshev's inequality therefore gives, with probability at least $1-\delta/2$,
\begin{equation}\label{eq:p2-convex-md-proof}
    \frac1N\left|\sum_{k=1}^N d_k\right| \le\frac{\sqrt2D\sigma_2}{\sqrt{N\delta}}.
\end{equation}

Also,
\[
    \E\left[\frac1N\sum_{k=1}^N\norm{g_k}^2\right] \le G_2^2.
\]
By Markov's inequality, with probability at least $1-\delta/2$,
\begin{equation}\label{eq:p2-convex-energy-proof}
    \frac1N\sum_{k=1}^N\norm{g_k}^2 \le\frac{2G_2^2}{\delta}.
\end{equation}
On the intersection of these two events,
\eqref{eq:generic-convex-sum} gives
\[
    F(\bar x_N)-F(x^\star) \le \frac{R_0^2}{2\eta N} +\frac{\eta G_2^2}{\delta} +\frac{\sqrt2D\sigma_2}{\sqrt{N\delta}}.
\]
With $\eta=R_0\sqrt\delta/(G_2\sqrt N)$, the first two terms sum to $3R_0G_2/(2\sqrt{N\delta})$.
Thus, with probability at least $1-\delta$,
\[
    F(\bar x_N)-F(x^\star) \le C\frac{R_0G_2+D\sigma_2}{\sqrt{N\delta}},
\]
which proves \eqref{eq:convex-p2-optimal}.
\end{proof}

\subsection{Proof of \eqref{eq:convex-p2-clipped}}
\label{proof_convex-p2-clipped}

\begin{proof}
The preceding clipping argument applies with $p=2$ and the raw second-moment bound $G_2^2$. In addition to \eqref{eq:clip-bias-variance}, we have
\[
    \E_{k-1}\norm{\widetilde\zeta_k(\lambda)}^2 \le\E_{k-1}\norm{h_k}^2 \le G_2^2.
\]
Thus the martingale bound can use $G_2$ as its variance scale, regardless of the stated value of $\sigma_2$.

Consequently, \eqref{eq:convex-clipped-general} holds with $p=2$, with $G_2^2/\lambda$ as the bias bound, and with $G_2$ in place of $\sigma_2$.
Choose
\[
    \eta=\frac{R_0}{G_2\sqrt N}, \quad \lambda=cG_2\sqrt{\frac{N}{\ell_\delta}},
\]
where $c>0$ is a fixed numerical constant. Then
\[
    D\lambda\frac{\ell_\delta}{N} +\frac{DG_2^2}{\lambda} =(c+c^{-1})DG_2\sqrt{\frac{\ell_\delta}{N}},
\]
and
\[
    \eta G_2\lambda\sqrt{\frac{\ell_\delta}{N}} =c\frac{R_0G_2}{\sqrt N}, \quad
    \eta\lambda^2\frac{\ell_\delta}{N} =c^2\frac{R_0G_2}{\sqrt N}.
\]
Substituting these identities into the general bound gives, with probability at least $1-\delta$,
\[
    F(\bar x_N)-F(x^\star) \le C\frac{R_0G_2}{\sqrt N} + CDG_2\sqrt{\frac{\ell_\delta}{N}},
\]
which proves \eqref{eq:convex-p2-clipped}.
\end{proof}

\subsection{Remainder conditions and sample requirements}
\label{proof_convex-comparison}

We verify the conditions used in the comparison in Section~\ref{sec:convex}. For $p>2$, write
\[
    r:=\frac12-\frac1p>0, \quad
    T_{\rm U}:= D\sigma_p\delta^{-1/p}N^{-(1-1/p)}, \quad T_{\rm C}:= DG_p\ell_\delta^{1-1/p}N^{-(1-1/p)}.
\]
These are the heavy-tail correction scales, with constants depending only on $p$ suppressed.

The polynomial squared-gradient remainders factor as
\begin{align}
    \frac{R_0\sigma_p^2}{G_2} \delta^{-2/p}N^{-(3/2-2/p)} &= T_{\rm U}
    \left( \frac{R_0\sigma_p}{DG_2} \delta^{-1/p}N^{-r} \right),\\
    \frac{R_0G_p^2}{G_2} \ell_\delta^{1-2/p}N^{-(3/2-2/p)}
    &=T_{\rm C}\left( \frac{R_0G_p}{DG_2} \ell_\delta^{-1/p}N^{-r} \right).
\end{align}
Thus the first condition in \eqref{eq:convex-remainder-regime} absorbs these two remainders into their respective heavy-tail corrections.

For $p>4$, the Gaussian component of $\frac{R_0}{2G_2\sqrt N}\mathcal Q_{p,N}(\delta)$ has scale
\[
    \frac{R_0\sigma_p^2}{G_2} \frac{\sqrt{\ell_\delta}}{N} = T_{\rm U} \left( \frac{R_0\sigma_p}{DG_2} \delta^{1/p}\ell_\delta^{1/2}N^{-1/p} \right).
\]
The second condition in \eqref{eq:convex-remainder-regime} absorbs this term as well.

After these absorptions, the ratio of the two correction scales is
\[
    \frac{T_{\rm U}}{T_{\rm C}}=\frac{\sigma_p}{G_p}\delta^{-1/p}\ell_\delta^{-(1-1/p)}.
\]
Taking the $p$-th power gives\eqref{eq:convex-usgd-better}, with constants depending only on $p$.

To obtain the sample requirements, observe that
\[
    N\ge K_pN_0(\eps,\delta)
\]
makes the optimization and Gaussian terms at most a prescribed constant fraction of $\eps$, for a sufficiently large $K_p$. Similarly,
\[
    N\ge K_p\left(\frac{D\sigma_p}{\eps}\right)^{p/(p-1)} \delta^{-1/(p-1)}
\]
controls the U-SGD correction, while
\[
    N\ge K_p\left(\frac{DG_p}{\eps}\right)^{p/(p-1)} \ell_\delta
\]
controls the WG-Clip correction.

Choosing each horizon as the ceiling of a sufficiently large $p$-dependent constant times the corresponding sum gives \eqref{eq:convex-complexities}, provided that the selected horizons also satisfy \eqref{eq:convex-remainder-regime}. The term $1$ in $N_0$ accounts for integer rounding. This is the stated conditional interpretation of those sample requirements; without the remainder conditions, the full bounds \eqref{eq:convex-unclipped-tuned} and \eqref{eq:convex-clipped-tuned} must be used.

Finally, solving \eqref{eq:convex-p2-optimal} for $N$ gives a sufficient budget of order
\[
    1+\left(\frac{R_0G_2+D\sigma_2}{\eps}\right)^2\delta^{-1}.
\]
Controlling the two terms in \eqref{eq:convex-p2-clipped} separately gives a sufficient budget of order
\[
    1+\left(\frac{R_0G_2}{\eps}\right)^2 +\left(\frac{DG_2}{\eps}\right)^2\ell_\delta.
\]
These are exactly \eqref{eq:convex-p2-complexities}.

\section{Epoch-Averaged SGD on Strongly Convex Quadratics}
\label{app:quadratic-epochs}
\label{sec:strong}

This appendix gives a finite-sample analysis of an explicit, unprojected epoch scheme for strongly convex quadratic objectives.  We use $\ell_\delta : =\log(4/\delta)$ for $0<\delta<1/2$.

\subsection{Quadratic model and oracle}
\label{subsec:quad-model}

We minimize an unconstrained quadratic on $\mathbb R^d$:
\begin{equation}
    F(x):=F^\star+\frac12(x-x^\star)^\top H(x-x^\star), \quad H=H^\top,\quad 0<\mu I\preceq H\preceq LI,\quad \kappa=\frac L\mu,
\end{equation}
where $H$ is deterministic and $F^\star=F(x^\star)$. In particular, $\nabla F(x)=H(x-x^\star)$ and $x^\star$ is the unique minimizer.

The smoothness and strong-convexity constants satisfy, for all points $x,y$ in the domain,
\begin{align}
    & \norm{\nabla F(x)-\nabla F(y)}\le L\norm{x-y},
    \\& F(y)\ge F(x)+\ip{\nabla F(x)}{y-x}+\frac{\mu}{2}\norm{y-x}^2.
\end{align}
We write $\kappa=L/\mu$ and $\Delta_0=F(x_1)-F(x^\star)$.

Within an epoch, let $\mathcal F_0$ contain all information available at its start. The starting point $x_0=z$ is $\mathcal F_0$-measurable. The filtration $\mathcal F_t$ includes the oracle response at step $t$, and $x_{t-1}$ is $\mathcal F_{t-1}$-measurable. For a fixed $p\ge2$, use the notation of Assumption~\ref{ass:noise} and write
\begin{equation}
    g_t=H(x_{t-1}-x^\star)+\zeta_t, \quad \mathbb E[\zeta_t\mid\mathcal F_{t-1}]=0,
\end{equation}
where, almost surely at every step,
\begin{equation}
    \mathbb E[\|\zeta_t\|^2\mid\mathcal F_{t-1}]\le\sigma_2^2, \quad \mathbb E[\|\zeta_t\|^p\mid\mathcal F_{t-1}]\le\sigma_p^p.
\end{equation}

The noise may depend on the current iterate and need not be independent across steps. The conditional bounds are assumed to hold along the entire unprojected run. Write the supplied direction as
\begin{equation}\label{eq:quad-direction}
    h_t=H(x_{t-1}-x^\star)+u_t. 
\end{equation}
Ordinary, unclipped SGD (U-SGD) uses $h_t=g_t$, hence $u_t=\zeta_t$.

The noise-centered oracle benchmark uses $u_t=\operatorname{clip}_{\lambda}(\zeta_t)$, leaving the exact mean gradient outside the clipping map. This benchmark generally requires information beyond one stochastic-gradient observation.

\begin{assumption}[Certified centered input]
\label{ass:certified-centered-input}
At every step of an epoch, the perturbation in \eqref{eq:quad-direction} has a decomposition
\begin{equation}\label{eq:certified-input-decomposition}
    u_t=v_t+b_t,\qquad b_t\text{ is }\mathcal F_{t-1}\text{-measurable},
\end{equation}
with deterministic epoch-wise bounds $\bar\sigma_2\ge0$, $\lambda>0$, and $B_\lambda\ge0$ such that, almost surely,
\begin{align}
    \mathbb E[v_t\mid\mathcal F_{t-1}]&=0,
    &\mathbb E[\|v_t\|^2\mid\mathcal F_{t-1}]&\le\bar\sigma_2^2,
    \\
    \|v_t\| & \le2\lambda,
    &\|b_t\|&\le B_\lambda.
\end{align}
A $p$-moment certificate with scale $M_p>0$ additionally means
\begin{equation}\label{eq:certified-p-bias}
    \bar\sigma_2\le\sigma_2,\quad B_\lambda\le c_{\rm cert}\frac{M_p^p}{\lambda^{p-1}},
\end{equation}
where $c_{\rm cert}$ is fixed.
\end{assumption}

The term \emph{centered} refers to $v_t$, not to the full perturbation: $b_t$ is retained in all bounds. Any anchor error or error from clipping a gradient difference must be included in $b_t$. We verify the noise-centered certificate in Lemma~\ref{lem:quad-clip-certificate} and the difference-clipping certificate in Proposition~\ref{prop:dc-certificate}.

\subsection{Epoch algorithm and restarts}
\label{subsec:quad-algorithms}

\paragraph{One epoch.}
Given a starting point $z$, the length $m\ge\lceil4\kappa\rceil$, and a choice of directions $h_t$, set
\begin{equation}
    x_0=z,\qquad \eta=\frac1{2L},\quad b=\lfloor m/2\rfloor,\qquad n=m-b.
\end{equation}
Run the $m$ unprojected updates
\begin{equation}\label{eq:main-quadratic-epoch-recursion}
    x_t=x_{t-1}-\eta h_t =x_{t-1}-\eta\bigl(H(x_{t-1}-x^\star)+u_t\bigr), \quad t=1,\ldots,m,
\end{equation}
and return
\begin{equation}
    z^+ :=\frac1n\sum_{t=b+1}^{m}x_t.
\end{equation}
The first $b$ steps are a burn-in period. A clipping threshold is constant throughout this epoch.

\paragraph{Geometric restarts.}
Let $z_0$ satisfy $F(z_0)-F^\star\le\Delta_0$ almost surely, where $\Delta_0>0$ is a known deterministic upper bound. Fix $0<\varepsilon<\Delta_0$ and $0<\delta<1/2$, and define
\begin{equation}
    S :=\left\lceil\log_2\frac{\Delta_0}{\varepsilon}\right\rceil, \quad \Delta_s=2^{-s}\Delta_0,\quad s=0,\ldots,S.
\end{equation}
Allocate failure probabilities by
\begin{equation}\label{eq:restart-failure-allocation-app}
    \delta_s=\frac{6\delta}{\pi^2(s+1)^2}, \quad s=0,\ldots,S-1,\quad \sum_{s=0}^{S-1}\delta_s\le\delta.
\end{equation}
For later use, set
\begin{equation}
    \ell_S :=\log\left(\frac{2\pi^2S^2}{3\delta}\right) =\max_{0\le s<S}\ell_{\delta_s}.
\end{equation}
\begin{samepage}
For each $s=0,\ldots,S-1$, carry out the following three steps:
\begin{enumerate}
    \item Choose $m_s$ with $(\Delta,\delta)=(\Delta_s,\delta_s)$ and the appropriate input type.
    \item Start a new epoch at $x_{s,0}=z_s$ and use \eqref{eq:main-quadratic-epoch-recursion} with $b_s=\lfloor m_s/2\rfloor$ and $n_s=m_s-b_s$. For a certified $p$-moment input, use $\lambda_s=M_{p,s}(m_s/\ell_{\delta_s})^{1/p}$.
    \item Set $z_{s+1}=n_s^{-1}\sum_{t=b_s+1}^{m_s}x_{s,t}$.
\end{enumerate}
Return $\widehat x :=z_S$ after $N_{\rm core} :=\sum_{s=0}^{S-1}m_s$ core updates.
\end{samepage}
If $\varepsilon\ge\Delta_0$, return $z_0$ without an update.

\begin{lemma}[Filter identity and norm budgets]
\label{lem:quad-filter}
For the epoch above, let $e_t:=x_t-x^\star$, $A:=I-\eta H$, $q:=1-1/(2\kappa)$, and $\bar e :=n^{-1}\sum_{t=b+1}^{m}e_t$. For every realized input sequence $(u_t)_{t=1}^m$,
\begin{equation}
    H^{1/2}\bar e =d_m-\sum_{i=1}^{m}W_{i,m}u_i,
    \label{eq:quadratic-filter-representation}
\end{equation}
where $W_{i,m}$ are deterministic and $d_m$ is $\mathcal F_0$-measurable, with
\begin{equation}
    d_m:=\frac1n\sum_{t=b+1}^{m}H^{1/2}A^te_0,\quad W_{i,m}:=\frac{\eta}{n}H^{1/2} \sum_{t=\max\{b+1,i\}}^{m}A^{t-i}.
\end{equation}
For every $r\ge1$,
\begin{equation}
    \sum_{i=1}^{m}\|W_{i,m}\|_{\rm op}^{r} \le \frac{2}{\mu^{r/2}n^{r-1}}, \quad \max_i\|W_{i,m}\|_{\rm op}\le\frac1{\sqrt\mu\,n}.
\end{equation}
In particular,
\begin{align}
    \sum_i\|W_{i,m}\|_{\rm op}^2&\le\frac4{\mu m},
    &\sum_i\|W_{i,m}\|_{\rm op}^p
    &\le\frac{2^p}{\mu^{p/2}m^{p-1}},
    \notag\\
    \max_i\|W_{i,m}\|_{\rm op}&\le\frac2{\sqrt\mu\,m},
    &\sum_i\|W_{i,m}\|_{\rm op}&\le\frac2{\sqrt\mu}.
    \label{eq:pr-filter-budgets}
\end{align}
Consequently, on $\{F(z)-F^\star\le\Delta\}$,
\begin{equation}
    F(z^+)-F^\star \le 2\Delta e^{-m/(2\kappa)} +\left\|\sum_{i=1}^{m}W_{i,m}u_i\right\|^2.
    \label{eq:pr-stable-reduction}
\end{equation}
\end{lemma}

\begin{proof}
The recursion is $e_t=Ae_{t-1}-\eta u_t$, and induction gives $e_t=A^te_0-\eta\sum_{i=1}^tA^{t-i}u_i$. Average this equality over $t=b+1,\ldots,m$ and interchange the finite sums to obtain \eqref{eq:quadratic-filter-representation}. Since $A$ is a polynomial in $H$, these matrices commute, and
\[
    \eta\sum_{j=0}^{r-1}A^j=H^{-1}(I-A^r),\qquad r\in\mathbb N.
\]
Hence
\[
    W_{i,m}=
    \begin{cases}
    n^{-1}H^{-1/2}A^{b+1-i}(I-A^n),&1\le i\le b,\\
    n^{-1}H^{-1/2}(I-A^{m-i+1}),&b+1\le i\le m.
    \end{cases}
\]
All eigenvalues of $A$ lie in $[1/2,q]$. Thus $\|I-A^r\|_{\rm op}\le1$ and
\[
    \|W_{i,m}\|_{\rm op}\le \frac1{\sqrt\mu\,n}
    \begin{cases}
    q^{b+1-i},&i\le b,\\
    1,&i>b.
    \end{cases}
\]
Because $n\ge m/2\ge2\kappa$ and, for $r\ge1$, $(1-q^r)^{-1}\le(1-q)^{-1}=2\kappa$, we obtain
\[
    \sum_i\|W_{i,m}\|_{\rm op}^r \le\frac{n+\sum_{j=1}^{b}q^{rj}}{\mu^{r/2}n^r} \le\frac{n+2\kappa}{\mu^{r/2}n^r} \le\frac2{\mu^{r/2}n^{r-1}}.
\]

Commutation also yields $\|d_m\|\le q^{b+1}\|H^{1/2}e_0\|\le q^{b+1}\sqrt{2\Delta}$. Using $2(b+1)\ge m$ and $q\le e^{-1/(2\kappa)}$, we have $\|d_m\|^2\le2\Delta e^{-m/(2\kappa)}$. Finally, $F(z^+)-F^\star=\frac12\|H^{1/2}\bar e\|^2$ and $\frac12\|a-b\|^2\le\|a\|^2+\|b\|^2$ give \eqref{eq:pr-stable-reduction}.
\end{proof}

\subsection{Concentration and clipping}
\label{subsec:quad-concentration}
If $(d_t)$ is a martingale-difference sequence and the deterministic budgets satisfy $\sum_t\mathbb E[\|d_t\|^2\mid\mathcal F_{t-1}]\le V$ and $\sum_t\mathbb E[\|d_t\|^p\mid\mathcal F_{t-1}]\le A_p$ almost surely, then, for $p>2$, the Fuk--Nagaev inequality gives
\begin{equation}
    \left\|\sum_td_t\right\| \le C\sqrt{V\ell_\delta}+C_p(A_p/\delta)^{1/p}
    \label{eq:quad-fn-input}
\end{equation}
with probability at least $1-\delta$; see \cite[Theorem~2.2]{MollenhauerFiedler2025}. If instead $\|d_t\|\le R$ almost surely, the Hilbert-space Bennett--Freedman inequality gives
\begin{equation}
    \left\|\sum_td_t\right\| \le C\sqrt{V\ell_\delta}+CR\ell_\delta
    \label{eq:quad-freedman-input}
\end{equation}
with the same probability; see \cite[Theorem~3.4]{Pinelis1994}. These inequalities also apply under the conditional law given $\mathcal F_0$ when the same deterministic budgets hold.

\begin{lemma}[Operator-weighted bounds with deterministic budgets]
\label{lem:operator-weighted-nagaev}
Let $W_t$ be deterministic linear maps between Euclidean spaces, and put $\mathcal W_r :=\sum_t\|W_t\|_{\rm op}^r$ and $\mathcal W_\infty :=\max_t\|W_t\|_{\rm op}$. Under Assumption~\ref{ass:noise}, for $p>2$,
\begin{equation}
    \left\|\sum_tW_t\zeta_t\right\| \le C\sigma_2\sqrt{\mathcal W_2\ell_\delta} +C_p\sigma_p \mathcal W_p^{1/p}\delta^{-1/p}
    \label{eq:operator-nagaev}
\end{equation}
with conditional probability at least $1-\delta$ given $\mathcal F_0$. For the centered component in
Assumption~\ref{ass:certified-centered-input},
\begin{equation}
    \left\|\sum_tW_tv_t\right\| \le C\bar\sigma_2\sqrt{\mathcal W_2\ell_\delta} +C\lambda \mathcal W_\infty\ell_\delta
    \label{eq:operator-freedman}
\end{equation}
with the same conditional probability. At $p=2$, the unclipped sum satisfies
\begin{equation}
    \mathbb E\left[\left\|\sum_tW_t\zeta_t\right\|^2 \,\middle|\,\mathcal F_0\right]\le\sigma_2^2\mathcal W_2.
    \label{eq:quad-weighted-second-moment}
\end{equation}
\end{lemma}

\begin{proof}
For $d_t=W_t\zeta_t$, the conditional second- and $p$-moment budgets are bounded by $\sigma_2^2\mathcal W_2$ and $\sigma_p^p\mathcal W_p$. Apply \eqref{eq:quad-fn-input}. For $d_t=W_tv_t$, the second-moment budget is $\bar\sigma_2^2\mathcal W_2$ and $\|d_t\|\le2\lambda \mathcal W_\infty$; \eqref{eq:quad-freedman-input} gives the second claim. For the last claim, distinct martingale increments are orthogonal in conditional $L^2$. Expanding the squared norm therefore leaves only $\sum_t\mathbb E[\|W_t\zeta_t\|^2\mid\mathcal F_0]$.
\end{proof}

\begin{remark}
All $\mathcal W_r$ are deterministic. More generally, predictable random maps can be used if deterministic almost-sure upper bounds for their norm sums and maximum replace $\mathcal W_r$ and $\mathcal W_\infty$.
\end{remark}

\begin{lemma}[Bias, variance, and the noise-centered certificate]
\label{lem:quad-clip-certificate}
For a random vector $Y$, a sigma-field $\mathcal G$, and $a:=\mathbb E[Y\mid\mathcal G]$, suppose $\mathbb E[\|Y\|^p\mid\mathcal G]\le D^p$. Then
\begin{align}
    \left\|\mathbb E[\operatorname{clip}_\lambda(Y) \mid\mathcal G]-a\right\|
    &\le D^p/\lambda^{p-1}, \label{eq:quad-clip-bias}\\
    \mathbb E\!\left[ \left\|\operatorname{clip}_\lambda(Y) -\mathbb E[\operatorname{clip}_\lambda(Y)\mid\mathcal G]\right\|^2 \,\middle|\,\mathcal G\right]
    &\le \mathbb E[\|Y-a\|^2\mid\mathcal G].
    \label{eq:quad-clip-variance}
\end{align}
Consequently, noise-centered clipping is certified with $\bar\sigma_2=\sigma_2$, $M_p=\sigma_p$, and $c_{\rm cert}=1$.
\end{lemma}

\begin{proof}
Pointwise,
\[
    \|\operatorname{clip}_\lambda(y)-y\| =(\|y\|-\lambda)_+ \le\|y\|\mathbf1_{\{\|y\|>\lambda\}} \le\|y\|^p/\lambda^{p-1}.
\]
Conditional expectation and Jensen's inequality give \eqref{eq:quad-clip-bias}. The clipping map is Euclidean projection onto a closed ball, hence is nonexpansive. Conditional expectation minimizes squared error, so the left-hand side of \eqref{eq:quad-clip-variance} is at most
\[
    \mathbb E[\|\operatorname{clip}_\lambda(Y) -\operatorname{clip}_\lambda(a)\|^2\mid\mathcal G] \le\mathbb E[\|Y-a\|^2\mid\mathcal G].
\]
Apply these facts with $Y=\zeta_t$ and $\mathcal G=\mathcal F_{t-1}$. Take $b_t=\mathbb E[\operatorname{clip}_\lambda(\zeta_t) \mid\mathcal F_{t-1}]$ and $v_t=\operatorname{clip}_\lambda(\zeta_t)-b_t$. Then $\|v_t\|\le2\lambda$ and all the stated certificate bounds follow.
\end{proof}

\begin{lemma}[One epoch for a quadratic objective]
\label{lem:one-epoch}
Let $m\ge\lceil4\kappa\rceil$, and suppose $F(z)-F^\star\le\Delta$. Each bound below holds with conditional probability at least $1-\delta$ given the information at the epoch start. For U-SGD satisfying Assumption~\ref{ass:noise} with $p>2$,
\begin{equation}
    F(z^+)-F^\star \le 2\Delta e^{-m/(2\kappa)} +\frac{C\sigma_2^2\ell_\delta}{\mu m}+\frac{C_p\sigma_p^2}{\mu} \delta^{-2/p}m^{-2(1-1/p)}.
    \label{eq:epoch-unclipped}
\end{equation}
For any certified centered input,
\begin{equation}
    F(z^+)-F^\star \le 2\Delta e^{-m/(2\kappa)} +\frac{C\bar\sigma_2^2\ell_\delta}{\mu m} +\frac C\mu \left(\frac{\lambda\ell_\delta}{m}+B_\lambda\right)^2.
    \label{eq:epoch-certified-general}
\end{equation}
If this input has a $p$-moment certificate, $p\ge2$, choose
\begin{equation}
    \lambda=M_p\left(\frac m{\ell_\delta}\right)^{1/p}.
    \label{eq:quad-threshold}
\end{equation}
Then
\begin{equation}
    F(z^+)-F^\star \le 2\Delta e^{-m/(2\kappa)} +\frac{C\sigma_2^2\ell_\delta}{\mu m}  +\frac{C_pM_p^2}{\mu} \left(\frac{\ell_\delta}{m}\right)^{2(1-1/p)}.
    \label{eq:epoch-clipped}
\end{equation}
For a random starting point, the claims apply on the $\mathcal F_0$-measurable event $\{F(z)-F^\star\le\Delta\}$.
\end{lemma}

\begin{proof}
For U-SGD, Lemma~\ref{lem:operator-weighted-nagaev} and \eqref{eq:pr-filter-budgets} imply
\[
    \left\|\sum_{t=1}^{m}W_{t,m}\zeta_t\right\| \le\frac{C\sigma_2}{\sqrt\mu} \sqrt{\frac{\ell_\delta}{m}}+\frac{C_p\sigma_p}{\sqrt\mu}\delta^{-1/p}m^{-(1-1/p)}.
\]
Square this bound using $(a+b)^2\le2a^2+2b^2$ and insert it into \eqref{eq:pr-stable-reduction}. For a certified input, apply \eqref{eq:operator-freedman} to the centered part and bound the bias pathwise:
\[
    \left\|\sum_tW_{t,m}b_t\right\| \le B_\lambda\sum_t\|W_{t,m}\|_{\rm op} \le\frac{2B_\lambda}{\sqrt\mu}.
\]
With conditional probability at least $1-\delta$, their sum is bounded by
\[
    \left\|\sum_tW_{t,m}(v_t+b_t)\right\| \le\frac{C\bar\sigma_2}{\sqrt\mu} \sqrt{\frac{\ell_\delta}{m}} +\frac{C\lambda\ell_\delta}{\sqrt\mu\,m} +\frac{2B_\lambda}{\sqrt\mu}.
\]
Squaring proves \eqref{eq:epoch-certified-general}. For \eqref{eq:quad-threshold}, both $\lambda\ell_\delta/m$ and $M_p^p/\lambda^{p-1}$ equal $M_p(\ell_\delta/m)^{1-1/p}$. Use \eqref{eq:certified-p-bias} and $\bar\sigma_2\le\sigma_2$ to obtain \eqref{eq:epoch-clipped}.
\end{proof}

\begin{proposition}[The finite-variance endpoint]
\label{prop:strong-p2-endpoint}
Under the same quadratic setup with $p=2$, U-SGD satisfies
\begin{equation}
    F(z^+)-F^\star \le 2\Delta e^{-m/(2\kappa)} +\frac{C\sigma_2^2}{\mu m\delta}
    \label{eq:epoch-unclipped-p2}
\end{equation}
with conditional probability at least $1-\delta$. A certified second-moment input with
\begin{equation}
    \lambda=M_2\sqrt{\frac m{\ell_\delta}}
    \label{eq:p2-strong-threshold}
\end{equation}
satisfies, with the same conditional probability,
\begin{equation}
    F(z^+)-F^\star \le 2\Delta e^{-m/(2\kappa)} +\frac{C(\sigma_2^2+M_2^2)\ell_\delta}{\mu m}.
    \label{eq:epoch-clipped-p2}
\end{equation}
\end{proposition}

\begin{proof}
Equations \eqref{eq:quad-weighted-second-moment} and \eqref{eq:pr-filter-budgets} give $\mathbb E[\|\sum_tW_{t,m}\zeta_t\|^2\mid\mathcal F_0] \le4\sigma_2^2/(\mu m)$. Conditional Markov's inequality bounds this squared norm by $4\sigma_2^2/(\mu m\delta)$ outside an event of conditional probability at most $\delta$. Apply \eqref{eq:pr-stable-reduction}. For the certified input, use the $p=2$ case of \eqref{eq:epoch-clipped}, whose proof uses bounded-increment concentration and does not require $p>2$.
\end{proof}

\subsection{Epoch lengths and restart complexity}

We now specify the epoch lengths and total update count.

\begin{proposition}[A deterministic epoch schedule that halves the gap]
\label{prop:quad-halving}
Let $0<\delta<1/2$ and $\Delta>0$. For $p>2$, set $\alpha=p/[2(p-1)]$. There are sufficiently large constants $K_0$ and $K_p$, independent of $\Delta,\delta$, such that the following choices ensure $F(z^+)-F^\star\le\Delta/2$ with conditional probability at least $1-\delta$ whenever $F(z)-F^\star\le\Delta$:
\begin{align}
    m_{\rm U}(\Delta,\delta)
    &=\left\lceil K_0\kappa+ K_0\frac{\sigma_2^2\ell_\delta}{\mu\Delta} +K_p\left(\frac{\sigma_p^2}{\mu\Delta}\right)^\alpha \delta^{-1/(p-1)} \right\rceil,
    \label{eq:epoch-complexity-U}\\
    m_{\rm C}(\Delta,\delta;M_p)
    &=\left\lceil K_0\kappa+ K_0\frac{\sigma_2^2\ell_\delta}{\mu\Delta} +K_p\left(\frac{M_p^2}{\mu\Delta}\right)^\alpha\ell_\delta \right\rceil.
    \label{eq:epoch-complexity-C}
\end{align}
The clipped choice uses \eqref{eq:quad-threshold}; its constants may depend on $c_{\rm cert}$. At $p=2$, suitable choices are
\begin{align}
    m_{\rm U}^{(2)}(\Delta,\delta)
    &=\left\lceil K_0\kappa+K_0\frac{\sigma_2^2}{\mu\Delta\delta}\right\rceil,
    \label{eq:quad-epoch-length-u2}\\
    m_{\rm C}^{(2)}(\Delta,\delta;M_2)
    &=\left\lceil K_0\kappa+ K_0\frac{(\sigma_2^2+M_2^2)\ell_\delta} {\mu\Delta}\right\rceil,
    \label{eq:quad-epoch-length-c2}
\end{align}
with \eqref{eq:p2-strong-threshold} in the second case.
\end{proposition}

\begin{proof}
Choose $K_0$ so that $m\ge K_0\kappa$ implies both $m\ge\lceil4\kappa\rceil$ and $2\Delta e^{-m/(2\kappa)}\le\Delta/6$. The Gaussian term in \eqref{eq:epoch-unclipped} is at most $\Delta/6$ when $m\ge C\sigma_2^2\ell_\delta/(\mu\Delta)$. Solving the remaining inequality gives
\[
    \frac{C_p\sigma_p^2}{\mu}\delta^{-2/p}m^{-2(p-1)/p} \le\frac\Delta6 \quad\text{if}\quad m\ge C_p \left(\frac{\sigma_p^2}{\mu\Delta}\right)^{p/[2(p-1)]} \delta^{-1/(p-1)}.
\]
These three bounds prove the U-SGD claim. Solving the last term of \eqref{eq:epoch-clipped} instead gives $m\ge C_p(M_p^2/(\mu\Delta))^\alpha\ell_\delta$. At $p=2$, apply the same reasoning to \eqref{eq:epoch-unclipped-p2} and \eqref{eq:epoch-clipped-p2}. The ceiling and the sum of the required lengths preserve each inequality.
\end{proof}

\begin{theorem}[Restarted quadratic scheme, with explicit restart factors]
\label{thm:strong-complexities}
Run the algorithm of Section~\ref{subsec:quad-algorithms}, using the epoch lengths of Proposition~\ref{prop:quad-halving}. Suppose the conditional noise bounds, or the required certified-input bounds, hold uniformly across the run. For $p>2$, the output of either corresponding scheme satisfies
\begin{equation}
    \mathbb P\bigl(F(\widehat x)-F^\star\le\varepsilon\bigr) \ge1-\delta.
\end{equation}
With $\alpha:=p/[2(p-1)]$, the U-SGD update count obeys
\begin{equation}
    N_{\rm U} \le C\kappa S +\frac{C\sigma_2^2}{\mu\varepsilon}\ell_S +C_p\left(\frac{\sigma_p^2}{\mu\varepsilon}\right)^\alpha  \delta^{-1/(p-1)}S^{2/(p-1)}.
    \label{eq:strong-U-complexity}
\end{equation}
For certified inputs with $M_{p,s}\le\overline M_p$ and uniform $c_{\rm cert}$, the core-update count obeys
\begin{equation}
    N_{\rm core,C} \le C\kappa S +\frac{C\sigma_2^2}{\mu\varepsilon}\ell_S +C_p\left(\frac{\overline M_p^2}{\mu\varepsilon}\right)^\alpha \ell_S.
    \label{eq:strong-C-complexity}
\end{equation}
Noise-centered clipping has $\overline M_p=\sigma_p$. For difference clipping, \eqref{eq:strong-C-complexity} counts only the core updates; the total oracle count is \eqref{eq:dc-total-budget}.
\end{theorem}

\begin{proof}
Let $G_s$ be the event that $F(z_j)-F^\star\le\Delta_j$ for every $j\le s$. It is measurable at the start of epoch $s$. The conditional halving guarantee gives
\[
    \mathbb P\bigl(G_s\cap\{F(z_{s+1})-F^\star>\Delta_{s+1}\}\bigr) \le\delta_s\,\mathbb P(G_s)\le\delta_s.
\]
Since $G_0$ holds almost surely, a union bound over the first failed epoch proves $\mathbb P(G_S)\ge1-\sum_s\delta_s\ge1-\delta$. On $G_S$, the final error is at most $\Delta_S\le\varepsilon$. No independence between epochs is required.

For every $r>0$, the geometric targets satisfy
\begin{equation}
    \sum_{s=0}^{S-1}\Delta_s^{-r} \le \frac{\varepsilon^{-r}}{1-2^{-r}}.
    \label{eq:quad-restart-geometric-sum}
\end{equation}
Indeed, $\Delta_{S-1}>\varepsilon$ and the earlier inverse powers decrease by the factor $2^{-r}$. Also, $\ell_{\delta_s}\le\ell_S$ and
\[
    \delta_s^{-1/(p-1)} \le\left(\frac{\pi^2}{6\delta}\right)^{1/(p-1)} S^{2/(p-1)}.
\]
Sum \eqref{eq:epoch-complexity-U}, applying \eqref{eq:quad-restart-geometric-sum} with $r=1$ and $r=\alpha$. The ceilings add at most $S$, which is absorbed by $C\kappa S$ because $\kappa\ge1$. This proves \eqref{eq:strong-U-complexity}. For \eqref{eq:strong-C-complexity}, sum \eqref{eq:epoch-complexity-C}, use $M_{p,s}\le\overline M_p$, and apply the same geometric sums.
\end{proof}

\begin{corollary}[Restart complexity under finite variance]
\label{cor:quad-restarts-p2}
At $p=2$, the same success guarantee holds with the schedules \eqref{eq:quad-epoch-length-u2}--\eqref{eq:quad-epoch-length-c2}, and
\begin{align}
    N_{\rm U}^{(2)}
    &\le C\kappa S+\frac{C\sigma_2^2S^2}{\mu\varepsilon\delta}, \label{eq:restart-u-p2}\\
    N_{\rm core,C}^{(2)}
    &\le C\kappa S+ \frac{C(\sigma_2^2+\overline M_2^2)\ell_S}{\mu\varepsilon}.
    \label{eq:restart-c-p2}
\end{align}
\end{corollary}

\begin{proof}
The conditional success argument in the theorem applies without change. Sum the two $p=2$ epoch lengths, using $\delta_s^{-1}\le\pi^2S^2/(6\delta)$, $\ell_{\delta_s}\le\ell_S$, and $\sum_s\Delta_s^{-1}\le2/\varepsilon$.
\end{proof}

\begin{remark}[Interpretation of the comparison]
The factor $S^{2/(p-1)}$ in \eqref{eq:strong-U-complexity}, and its $p=2$ counterpart $S^2$, come from the chosen allocation \eqref{eq:restart-failure-allocation-app}. They are polynomial in the number of epochs $S$ and therefore polylogarithmic in $\Delta_0/\varepsilon$. For fixed $p>2$ and fixed confidence, $\alpha<1$, so the displayed heavy-tail terms have a lower order in $1/\varepsilon$ than the variance term, even with these restart factors. At $p=2$, both statistical terms have order $1/\varepsilon$, and the confidence dependence distinguishes the displayed bounds. These are upper bounds for the specified schemes, not a minimax comparison with all robust estimators or aggregations of multiple runs.
\end{remark}

\subsection{Difference clipping and its oracle cost}

We now consider the difference-clipped direction introduced in Section~\ref{sec:model}. We allow the anchor and its gradient estimate to depend on time, writing them as $\widetilde x_t$ and $\widetilde s_t$. The following proposition gives sufficient conditions on their accuracy and on the coupled gradient differences.

\begin{proposition}[Sufficient certificate for difference clipping]
\label{prop:dc-certificate}
Let $\widetilde x_t$ and $\widetilde s_t$ be $\mathcal F_{t-1}$-measurable. Suppose a coupled-sample oracle returns
\begin{equation}
    \begin{split}
    d_t&:=g(x_{t-1},\xi_t)-g(\widetilde x_t,\xi_t),\\
    a_t&:=\mathbb E[d_t\mid\mathcal F_{t-1}] =H(x_{t-1}-\widetilde x_t).
    \end{split}
\end{equation}
Use $h_t:=\widetilde s_t+\operatorname{clip}_\lambda(d_t)$. Assume deterministic epoch-wise bounds $a,S_2,D_p$ satisfy
\begin{align}
    \|\widetilde s_t-H(\widetilde x_t-x^\star)\|\le a,
    \quad
    \mathbb E[\|d_t-a_t\|^2\mid\mathcal F_{t-1}]\le S_2^2, \quad
    \mathbb E[\|d_t\|^p\mid\mathcal F_{t-1}]\le D_p^p
\end{align}
almost surely at every step. Then the input is certified with
\begin{equation}
    \bar\sigma_2=S_2,\quad B_\lambda=a+\frac{D_p^p}{\lambda^{p-1}}.
    \label{eq:dc-certificate-conclusion}
\end{equation}
In particular, a $p$-moment certificate with scale $M_p$ follows if $S_2\le\sigma_2$, $D_p\le c_D M_p$, and $a\le c_aM_p^p/\lambda^{p-1}$, with $c_{\rm cert}=c_a+c_D^p$ independent of the epoch and its parameters.
\end{proposition}

\begin{proof}
Define
\[
     \begin{split}
    v_t&=\operatorname{clip}_\lambda(d_t)-\mathbb E[\operatorname{clip}_\lambda(d_t)\mid\mathcal F_{t-1}],\\
    b_t&=\widetilde s_t-H(\widetilde x_t-x^\star) +\mathbb E[\operatorname{clip}_\lambda(d_t) \mid\mathcal F_{t-1}]-a_t.
    \end{split}
\]
Then $h_t-H(x_{t-1}-x^\star)=v_t+b_t$, $\mathbb E[v_t\mid\mathcal F_{t-1}]=0$, and $\|v_t\|\le2\lambda$. Apply Lemma~\ref{lem:quad-clip-certificate} to $Y=d_t$: the conditional centered variance is at most $S_2^2$, and the clipping bias is at most $D_p^p/\lambda^{p-1}$. Adding the anchor error gives \eqref{eq:dc-certificate-conclusion} and then the claimed $p$-moment certificate.
\end{proof}

\begin{remark}[Core updates and total oracle cost]
\label{rem:dc-budget}
The difference and anchor bounds above must be verified for the actual implementation, including each anchor refresh. Its total stochastic-gradient evaluation count is
\begin{equation}
    N_{\rm total}^{\rm DC} =N_{\rm core,C}+N_{\rm anchor}+N_{\rm refresh}+N_{\rm pair}.
    \label{eq:dc-total-budget}
\end{equation}
Here $N_{\rm anchor}$ counts initial anchor construction, $N_{\rm refresh}$ counts subsequent anchor work, and $N_{\rm pair}$ counts paired evaluations beyond the one charged to each core update. If a pair costs two evaluations, $N_{\rm pair}=N_{\rm core,C}$; it is zero only in an oracle model returning a coupled pair at unit cost. The bounds above control $N_{\rm core,C}$ after certification.
\end{remark}

\section{Polyak--Ruppert Asymptotics: Assumptions and Proofs}
\label{app:strong-proofs}

This appendix provides the complete statements and proofs underlying Section~\ref{sec:asymptotics}. Throughout, $\|\cdot\|$ denotes the Euclidean norm or its induced operator norm, as appropriate. We write $\|X\|_{L^2}:=(\mathbb E\|X\|^2)^{1/2}$ and $\ell_\delta:=\log(4/\delta)$.

\subsection{The exact linear identity and the unclipped limit}
Let $e_k:=x_k-x^\star$ and consider
\begin{equation}
    e_{k+1} :=e_k-\eta_k(He_k+\zeta_{k+1}), \quad H=H^\top\succ0,
    \label{eq:linear-sa}
\end{equation}
where $e_k$ is $\mathcal F_k$-measurable and $\mathbb E[\zeta_{k+1}\mid\mathcal F_k]=0$. The stepsizes are deterministic and positive. Define
\begin{equation}
    \bar e_N:=\frac1N\sum_{k=1}^N e_k, \quad \bar x_N:=\frac1N\sum_{k=1}^N x_k.
\end{equation}

\begin{proposition}[Exact Polyak--Ruppert decomposition]
\label{prop:exact-pr}
For every positive stepsize sequence,
\begin{equation}
    H\bar e_N =-\frac1N\sum_{k=1}^N\zeta_{k+1}+R_N,
    \label{eq:exact-pr}
\end{equation}
where
\begin{equation}
    R_N:=\frac1N\left[ \frac{e_1}{\eta_1}-\frac{e_{N+1}}{\eta_N}+\sum_{k=2}^Ne_k\left(\frac1{\eta_k}-\frac1{\eta_{k-1}}\right) \right].
    \label{eq:pr-remainder}
\end{equation}
The identity is pathwise and does not require symmetry or stability of the drift matrix.

Suppose, additionally, that $\eta_k:=a(k+k_0)^{-\gamma}$ with $a>0$, $k_0\ge0$, $\gamma\in(1/2,1)$, and $\|e_k\|_{L^2}\le C_e k^{-\gamma/2}$. Then
\begin{equation}
    \|R_N\|_{L^2} \le C N^{\gamma/2-1} =o(N^{-1/2}).
    \label{eq:pr-remainder-rate}
\end{equation}
\end{proposition}

\begin{proof}
Rearranging the update gives
\[
    He_k=\frac{e_k-e_{k+1}}{\eta_k}-\zeta_{k+1}.
\]
Summation by parts yields
\[
    \sum_{k=1}^N\frac{e_k-e_{k+1}}{\eta_k} = \frac{e_1}{\eta_1}-\frac{e_{N+1}}{\eta_N} +\sum_{k=2}^Ne_k\left(\frac1{\eta_k}-\frac1{\eta_{k-1}}\right),
\]
which proves \eqref{eq:exact-pr}--\eqref{eq:pr-remainder}.

For the stated stepsizes,
\[
    0\le \frac1{\eta_k}-\frac1{\eta_{k-1}} \le Ck^{\gamma-1}.
\]
Minkowski's inequality therefore gives
\[
    \|R_N\|_{L^2}\le\frac C N\left[1+N^{\gamma/2}+\sum_{k=2}^N k^{\gamma/2-1}\right] \le C N^{\gamma/2-1}.
\]
Since $\gamma<1$, this is $o(N^{-1/2})$.
\end{proof}

To obtain a Gaussian limit, suppose that the noise is square integrable and satisfies
\begin{equation}
    Q_N:= \frac1N\sum_{k=1}^N \mathbb E[\zeta_{k+1}\zeta_{k+1}^{\top}\mid\mathcal F_k] \xrightarrow{\mathbb P}\Sigma,
    \label{eq:martingale-conditional-covariance}
\end{equation}
where $\Sigma$ is deterministic, and, for every $\epsilon>0$,
\begin{equation}
    L_N(\epsilon):= \frac1N\sum_{k=1}^N
    \mathbb E\left[\|\zeta_{k+1}\|^2\mathbf1_{\{\|\zeta_{k+1}\|>\epsilon\sqrt N\}}\,\middle|\,\mathcal F_k \right] \xrightarrow{\mathbb P}0.
    \label{eq:martingale-lindeberg}
\end{equation}
Under the stepsize and moment-decay assumptions of Proposition~\ref{prop:exact-pr}, these conditions imply
\begin{equation}
    \sqrt N\,\bar e_N \Longrightarrow \mathcal N(0,H^{-1}\Sigma H^{-1}).
    \label{eq:classical-pr-clt}
\end{equation}
Indeed, the martingale central limit theorem \cite{HallHeyde1980} applies to $N^{-1/2}\sum_{k=1}^N\zeta_{k+1}$, while $\sqrt N R_N\to0$ in $L^2$. Equation~\eqref{eq:exact-pr} and Slutsky's theorem then give \eqref{eq:classical-pr-clt}, recovering the classical Polyak--Ruppert limit \cite{Ruppert1988,PolyakJuditsky1992}.

For i.i.d.\ centered noise independent of the initial condition, finite covariance implies both noise conditions above. Let an $\mathcal F_0$-measurable variable $U$ equal $1$ or $2$ with equal probability, and let $\zeta_k=U\varepsilon_k$, where the $\varepsilon_k$ are independent Rademacher variables independent of $U$. Then $Q_N=U^2$, and the normalized noise sum converges to
\[
    \tfrac12\mathcal N(0,1)+\tfrac12\mathcal N(0,4).
\]

If, in addition, for some $p>2$,
\[
    \mathbb E[\|\zeta_{k+1}\|^r\mid\mathcal F_k] \le\sigma_r^r, \quad r\in\{2,p\},
\]
uniformly in $k$, Theorem~\ref{thm:fuk-nagaev} and \eqref{eq:exact-pr} give, with probability at least $1-\delta$,
\begin{equation}
    \|\bar e_N\| \le \|H^{-1}\| \left[ C\sigma_2\sqrt{\frac{\ell_\delta}{N}}+C_p\sigma_p\delta^{-1/p}N^{-(1-1/p)} \right] +\|H^{-1}R_N\|.
\end{equation}
Thus the Gaussian limit and the finite-sample heavy-tail correction follow from the same exact representation.

\subsection{Fixed-threshold clipped stochastic approximation}

Fix $0<\lambda<\infty$ and define
\begin{equation}
    \psi_\lambda(x,\xi):=\clip_\lambda(g(x,\xi)),
    \quad
    h_\lambda(x):=\mathbb E\psi_\lambda(x,\xi).
\end{equation}
Consider
\begin{equation}
    x_{k+1} =x_k-\eta_k\psi_\lambda(x_k,\xi_{k+1}),
    \quad \bar x_N=\frac1N\sum_{k=1}^N x_k,
\end{equation}
where the samples are i.i.d.\ and independent of the initial condition. Its equilibrium is determined by $h_\lambda$.

\begin{assumption}[Fixed-threshold PR regularity]
\label{ass:fixed-clip-pr}
The following conditions hold.
\begin{enumerate}
\item The mean field $h_\lambda$ has a locally unique root $x_\lambda$. The matrix $A_\lambda:=\nabla h_\lambda(x_\lambda)$ has eigenvalues with positive real parts. For some $\alpha\in(0,1]$,
\begin{equation}
    \|h_\lambda(x_\lambda+y)-A_\lambda y\| \le C\|y\|^{1+\alpha}
    \label{eq:fixed-mean-remainder}
\end{equation}
for all sufficiently small $y$.

\item Define $M_\lambda(x,\xi):=\psi_\lambda(x,\xi)-h_\lambda(x)$. For some $\beta\in(0,1]$,
\begin{equation}
    \mathbb E \|M_\lambda(x_\lambda+y,\xi)-M_\lambda(x_\lambda,\xi)\|^2 \le C\|y\|^{2\beta}
    \label{eq:fixed-noise-continuity}
\end{equation}
for all sufficiently small $y$.

\item For $\eta_k=a(k+k_0)^{-\gamma}$ with $a>0$, $k_0\ge0$, and $\gamma\in(1/2,1)$, the recursion converges almost surely to $x_\lambda$. Writing $y_k:=x_k-x_\lambda$, assume
\begin{equation}
    \mathbb E\|y_k\|^2\le C\eta_k, \quad \mathbb E\|y_k\|^{2(1+\alpha)} \le C\eta_k^{1+\alpha}, \quad \gamma(1+\alpha)>1.
    \label{eq:fixed-stability-moments}
\end{equation}
\end{enumerate}
\end{assumption}

\begin{remark}
The three conditions separate local stability of the clipped mean field, continuity of the noise, and control of the actual iterates. They provide explicit sufficient conditions for replacing the recursion by its linearization around $x_\lambda$.

The first condition makes the linearized dynamics $\dot y=-A_\lambda y$ asymptotically stable and controls the error of this linear approximation. The remainder estimate holds, for example, when $h_\lambda$ has a locally $\alpha$-Hölder continuous Jacobian. A locally Lipschitz Jacobian gives $\alpha=1$. The required smoothness concerns the expected clipped score $h_\lambda$: individual clipped scores may be nonsmooth at the clipping boundary.

The second condition allows the state-dependent noise to be replaced, to first order, by the noise evaluated at $x_\lambda$. A simple sufficient condition for $\beta=1$ is local Lipschitz continuity of the stochastic gradient in mean square:
\[
    \mathbb E\|g(x_\lambda+y,\xi)-g(x_\lambda,\xi)\|^2  \le L_g^2\|y\|^2.
\]
Indeed, clipping is nonexpansive, and subtracting the expectation cannot increase the second moment of the score difference. This condition holds directly for additive linear models. Under the iterate moment bound in the third condition, every $\beta>0$ makes the accumulated noise-replacement error negligible on the root-$N$ scale.

The third condition controls convergence to the selected root and the size of the fluctuations around it. These properties require a separate stability argument for the recursion. The restriction $\gamma(1+\alpha)>1$ ensures that the accumulated nonlinear drift is negligible, since
\[
    \frac1{\sqrt N}\sum_{k=1}^N \mathbb E\|h_\lambda(x_\lambda+y_k)-A_\lambda y_k\| \lesssim \frac1{\sqrt N}\sum_{k=1}^N\eta_k^{(1+\alpha)/2} \longrightarrow0.
\]
For $\alpha=1$, this restriction reduces to $\gamma>1/2$; weaker regularity requires $\gamma>1/(1+\alpha)$. The upper restriction $\gamma<1$ makes the summation-by-parts remainder negligible on the same scale.

The higher-moment bound in \eqref{eq:fixed-stability-moments} is stronger than needed for the displayed CLT proof. Because $\alpha\le1$,
\[
    \mathbb E\|y_k\|^{1+\alpha} \le \bigl(\mathbb E\|y_k\|^2\bigr)^{(1+\alpha)/2} \lesssim \eta_k^{(1+\alpha)/2}.
\]
Thus the second-moment bound already supplies the required nonlinear-drift estimate. Finally, fixed clipping makes the score at the root bounded, so its Lindeberg condition is automatic.
\end{remark}

\begin{theorem}[Fixed-threshold clipped PR limit]
\label{thm:fixed-clip-clt}
Under Assumption~\ref{ass:fixed-clip-pr}, let
\begin{equation}
    \Sigma_\lambda :=\Cov\bigl(\psi_\lambda(x_\lambda,\xi)\bigr).
\end{equation}
Then
\begin{equation}
    \sqrt N(\bar x_N-x_\lambda) \Longrightarrow \mathcal N\left(= 0,A_\lambda^{-1}\Sigma_\lambda A_\lambda^{-\top} \right).
    \label{eq:fixed-clip-clt}
\end{equation}
If $x_\lambda\ne x^\star$, the sequence $\sqrt N(\bar x_N-x^\star)$ is not tight. In particular, $h_\lambda(x^\star)\ne0$ excludes a root-$N$ limit centered at the optimizer.
\end{theorem}

\begin{proof}
Define
\[
    \varepsilon_{k+1}:=M_\lambda(x_\lambda,\xi_{k+1}),
    \quad \Delta M_{k+1}:= M_\lambda(x_\lambda+y_k,\xi_{k+1})-M_\lambda(x_\lambda,\xi_{k+1}),
\]
and $r_h(y):=h_\lambda(x_\lambda+y)-A_\lambda y$. Then
\[
    y_{k+1} =y_k-\eta_k
    \bigl(A_\lambda y_k+\varepsilon_{k+1} +\Delta M_{k+1}+r_h(y_k)\bigr).
\]
Summation by parts gives
\[
    A_\lambda\bar y_N= -\frac1N\sum_{k=1}^N\varepsilon_{k+1} -\frac1N\sum_{k=1}^N\Delta M_{k+1} -\frac1N\sum_{k=1}^N r_h(y_k) +R_N,
\]
where $R_N$ is \eqref{eq:pr-remainder} with $e_k$ replaced by $y_k$. The moment bound in \eqref{eq:fixed-stability-moments} and the proof of Proposition~\ref{prop:exact-pr} imply $\sqrt N R_N\to0$ in $L^2$.

The local estimates \eqref{eq:fixed-mean-remainder}--\eqref{eq:fixed-noise-continuity} extend to all $y$ after increasing their constants. Indeed, $\|\psi_\lambda\|\le\lambda$ and $\|M_\lambda\|\le2\lambda$. Outside any fixed ball around zero,
\[
    \|r_h(y)\| \le\lambda+\|A_\lambda\|\,\|y\| \le C\|y\|^{1+\alpha},
\]
and the squared noise difference is bounded by $16\lambda^2\le C\|y\|^{2\beta}$.

Consequently,
\[
    \mathbb E \left\|\frac1{\sqrt N}\sum_{k=1}^N r_h(y_k)\right\| \le \frac C{\sqrt N}
    \sum_{k=1}^N\eta_k^{(1+\alpha)/2} \longrightarrow0,
\]
because $\gamma(1+\alpha)>1$.

The sequence $\Delta M_{k+1}$ is a martingale difference. Orthogonality of its increments and Jensen's inequality give
\[
    \mathbb E \left\|\frac1{\sqrt N}\sum_{k=1}^N\Delta M_{k+1}\right\|^2 \le \frac C N\sum_{k=1}^N\mathbb E\|y_k\|^{2\beta} \le \frac C N\sum_{k=1}^N\eta_k^\beta \longrightarrow0.
\]
It follows that
\[
    \sqrt N(\bar x_N-x_\lambda) = -A_\lambda^{-1}\frac1{\sqrt N}\sum_{k=1}^N\varepsilon_{k+1} +o_{\mathbb P}(1).
\]
The root scores $\varepsilon_{k+1}$ are i.i.d., centered, bounded, and have covariance $\Sigma_\lambda$. The ordinary multivariate CLT and Slutsky's theorem prove \eqref{eq:fixed-clip-clt}.

Finally,
\[
    \sqrt N(\bar x_N-x^\star) = \sqrt N(x_\lambda-x^\star) +\sqrt N(\bar x_N-x_\lambda).
\]
If $x_\lambda\ne x^\star$, the first term diverges in norm and the second is tight.
\end{proof}

The next subsection gives an exact quadratic example with $x_\lambda\ne x^\star$.

\subsection{A quadratic counterexample for fixed whole-gradient clipping}
\label{subsec:quad-whole-gradient-bias}

Whole-gradient clipping supplies $h_t=\operatorname{clip}_\lambda(H(x_{t-1}-x^\star)+\zeta_t)$. It clips the signal as well as the noise. 


The following exact quadratic example records the resulting possible shift of the target. The general stochastic clipping-bias phenomenon is established by \cite{KoloskovaHendrikxStich2023}.

\begin{proposition}[Shifted equilibrium under asymmetric noise]
\label{prop:asymmetric-bias}
Let
\begin{equation}
    F(x)=\frac{\mu x^2}{2},\qquad g(x,\xi)=\mu x+\xi,\qquad \xi=
    \begin{cases}
    a,&\text{with probability }q,\\
    -b,&\text{with probability }1-q,
    \end{cases}
    \quad b=\frac{qa}{1-q},
\end{equation}
where $a>0$ and $0<q<1/2$. Thus $\mathbb E\xi=0$ and $x^\star=0$. For a fixed threshold $b<\lambda<a$, the clipped mean field $h_\lambda(x)=\mathbb E[\operatorname{clip}_\lambda(\mu x+\xi)]$ has the unique root
\begin{equation}
    x_\lambda=\frac{q(a-\lambda)}{(1-q)\mu}>0.
    \label{eq:asym-root}
\end{equation}
For i.i.d.\ copies $\xi_k$, deterministic finite $x_0$, and deterministic steps $\eta_k>0$ with $\sum_k\eta_k=\infty$ and $\sum_k\eta_k^2<\infty$, the recursion
\begin{equation}
    x_{k+1}=x_k-\eta_{k+1} \operatorname{clip}_\lambda(\mu x_k+\xi_{k+1})
\end{equation}
satisfies $x_k\to x_\lambda$ almost surely. In particular, its arithmetic averages also converge to $x_\lambda$, and
\begin{equation}
    F(x_\lambda)-F(x^\star)=\frac{q^2(a-\lambda)^2}{2\mu(1-q)^2}>0.
    \label{eq:asym-gap}
\end{equation}
\end{proposition}

\begin{proof}
At the proposed root, $\mu x_\lambda-b=-q\lambda/(1-q)\in(-\lambda,0)$ and $\mu x_\lambda+a>\lambda$. In a neighborhood of this point,
\[
    h_\lambda(x)=q\lambda+(1-q)(\mu x-b)=q\lambda+(1-q)\mu x-qa.
\]
Its zero is \eqref{eq:asym-root}. The scalar clipping map is continuous and nondecreasing, so $h_\lambda$ has these properties as well. It is strictly increasing near its displayed root; monotonicity then implies that this root is unique and that $(x-x_\lambda)h_\lambda(x)>0$ for every $x\ne x_\lambda$.

Let $V_k=(x_k-x_\lambda)^2$ and $\phi(x)=(x-x_\lambda)h_\lambda(x)\ge0$. The direction is bounded by $\lambda$, and hence
\[
    \mathbb E[V_{k+1}\mid\mathcal F_k] \le V_k-2\eta_{k+1}\phi(x_k)+\eta_{k+1}^2\lambda^2.
\]
Consequently $V_k+\lambda^2\sum_{j=k+1}^{\infty}\eta_j^2$ is a nonnegative supermartingale and converges almost surely. The deterministic tail sum tends to zero, so $V_k$ has a finite almost-sure limit. Taking expectations and summing the drift inequality also shows $\sum_k\eta_{k+1}\phi(x_k)<\infty$ almost surely. If the limit of $V_k$ were positive, the iterates would eventually belong to a compact set bounded away from $x_\lambda$. Continuity and the strict sign property would then give $\phi(x_k)\ge c>0$ along the tail, contradicting $\sum_k\eta_k=\infty$. Thus $V_k\to0$, proving convergence; convergence of arithmetic averages follows. Substitution into $F$ gives \eqref{eq:asym-gap}.
\end{proof}

The shrinking-error bounds in Appendix~\ref{app:quadratic-epochs} use epoch-dependent thresholds and the stated certificates. The fixed whole-gradient recursion in this last proposition is a different policy, so its nonzero limiting error is consistent with those bounds.

\subsection{Symmetric noise and the covariance comparison}

\begin{corollary}[One-dimensional symmetric formula]
\label{cor:symmetric-variance}
Let $F(x)=\mu x^2/2$ with $\mu>0$ and $g(x,\xi)=\mu x+\xi$, where $\xi$ has a continuous distribution symmetric around zero and finite variance. Suppose $\mathbb P(|\xi|<\lambda)>0$ and Assumption~\ref{ass:fixed-clip-pr} holds at $x_\lambda=0$. Then $x_\lambda=x^\star=0$ and
\begin{equation}
    A_\lambda=\mu\mathbb P(|\xi|<\lambda),
    \quad
    \Sigma_\lambda=\mathbb E\min\{\xi^2,\lambda^2\}.
\end{equation}
The asymptotic variance of $\sqrt N\bar x_N$ is
\begin{equation}
    V_\lambda = \frac{\mathbb E\min\{\xi^2,\lambda^2\}}{\mu^2\mathbb P(|\xi|<\lambda)^2}.
    \label{eq:clipped-asymptotic-variance}
\end{equation}
\end{corollary}

\begin{proof}
Symmetry and oddness imply $h_\lambda(0)=\mathbb E\clip_\lambda(\xi)=0$. For $|\xi|\ne\lambda$,
\[
    \left.\frac{\partial}{\partial x} \clip_\lambda(\mu x+\xi)\right|_{x=0} = \mu\mathbf1_{\{|\xi|<\lambda\}}.
\]
The difference quotients are bounded by $\mu$, and continuity of the noise law excludes mass at $\pm\lambda$. Dominated convergence therefore gives the stated $A_\lambda$. The root score has mean zero and squared value $\min\{\xi^2,\lambda^2\}$, proving the formula for $\Sigma_\lambda$. Apply Theorem~\ref{thm:fixed-clip-clt}.
\end{proof}

\paragraph{Gaussian noise.}
Let $\xi\sim\mathcal N(0,\sigma^2)$ with $\sigma>0$. Gaussian integration by parts and Cauchy--Schwarz give
\begin{equation}
    \sigma^2\mathbb P(|\xi|<\lambda) = \mathbb E[\xi\clip_\lambda(\xi)] \le \sigma\left(\mathbb E\min\{\xi^2,\lambda^2\} \right)^{1/2}.
\end{equation}
Consequently, $V_\lambda\ge\sigma^2/\mu^2$. For every finite $\lambda>0$, the inequality is strict: equality would require $\clip_\lambda(\xi)$ to be proportional to $\xi$ almost surely, which fails for a nondegenerate Gaussian law. Dominated convergence gives $V_\lambda\to\sigma^2/\mu^2$ as $\lambda\to\infty$.

\paragraph{A symmetric heavy-tailed example.}
The reverse comparison is possible for other noise laws. If $\xi$ has a density $f$ continuous and positive at zero, then
\[
    \frac{\mathbb E\min\{\xi^2,\lambda^2\}}{\lambda^2}\to1,
    \quad
    \frac{\mathbb P(|\xi|<\lambda)}{\lambda}\to2f(0)
    \quad(\lambda\downarrow0).
\]
Thus $V_\lambda\to[4\mu^2 f(0)^2]^{-1}$. For a standard Student $t_3$ variable,
\[
    f(x)=\frac{2}{\pi\sqrt3}\left(1+\frac{x^2}{3}\right)^{-2},
    \quad \mathbb E\xi^2=3,
\]
and hence
\[
    \lim_{\lambda\downarrow0}V_\lambda =\frac{3\pi^2}{16\mu^2} <\frac3{\mu^2}.
\]
Therefore sufficiently small positive thresholds give a smaller variance in \eqref{eq:clipped-asymptotic-variance} than the unclipped value. The PR interpretation of this comparison uses the regularity and stability assumptions of Corollary~\ref{cor:symmetric-variance}.

\subsection{Growing thresholds in the linear additive model}

Let $H=H^\top\succ0$, and write $\mu=\lambda_{\min}(H)$ and $L=\lambda_{\max}(H)$. Let $(\zeta_k)$ be i.i.d.\ with
\[
    \mathbb E\zeta_k=0, \quad \mathbb E[\zeta_k\zeta_k^\top]=\Sigma, \quad\mathbb E\|\zeta_k\|^2\le\sigma_2^2, \quad \mathbb E\|\zeta_k\|^p\le\sigma_p^p,
\]
where $p>2$ and the moment bounds are finite and positive.

For each horizon $N$, use one deterministic threshold $\lambda_N>0$ throughout the recursion
\begin{equation}
    e_{k+1}^{(N)} = e_k^{(N)} -\eta_k\left( He_k^{(N)}+\clip_{\lambda_N}(\zeta_{k+1}) \right), \quad k=1,\ldots,N.
\end{equation}
Use the common initialization $e_1^{(N)}=e_1$, independent of the noise sequence, with $\mathbb E\|e_1\|^2<\infty$. Define
\begin{equation}
    b_N:=\mathbb E\clip_{\lambda_N}(\zeta),
    \quad
    e_N^\star:=-H^{-1}b_N,
    \quad
    v_{N,k}:=\clip_{\lambda_N}(\zeta_k)-b_N.
\end{equation}
The vector $e_N^\star$ is the exact equilibrium of the horizon-dependent mean field. Set
\[
    y_k^{(N)}:=e_k^{(N)}-e_N^\star, \quad \bar e_N^{(N)}:=\frac1N\sum_{k=1}^N e_k^{(N)}.
\]

\begin{theorem}[A sufficient growing-threshold window]
\label{thm:growing-window}
Let $\eta_k=a(k+k_0)^{-\gamma}$ with $a>0$, $k_0\ge0$, and $\gamma\in(1/2,1)$. Let $\delta_N\in(0,1/2)$ and $\ell_N:=\log(4/\delta_N)$.
Suppose
\begin{equation}
    \lambda_N\to\infty, \quad \frac{\sqrt N\,\sigma_p^p}{\lambda_N^{p-1}}\to0, \quad \frac{\lambda_N\ell_N}{\sqrt N}\to0.
    \label{eq:growing-rigorous-conditions}
\end{equation}
Then
\begin{equation}
    \sqrt N\,\bar e_N^{(N)} \Longrightarrow \mathcal N(0,H^{-1}\Sigma H^{-1}).
    \label{eq:growing-same-clt}
\end{equation}

Let $R_N^{(N)}$ be \eqref{eq:pr-remainder} applied to $(y_k^{(N)})$. There is a constant $C_R$, independent of $N$ and the thresholds, such that
\begin{equation}
    \|R_N^{(N)}\|_{L^2} \le C_R N^{\gamma/2-1}.
    \label{eq:growing-remainder-l2}
\end{equation}
For every $\rho_N\in(0,1/2)$, with probability at least $1-\delta_N-\rho_N$,
\begin{equation}\label{eq:growing-finite-bound}
\begin{split}
    \|\bar e_N^{(N)}\|
    \le \|H^{-1}\|\bigg[
    &\frac{\sigma_p^p}{\lambda_N^{p-1}} +C\sigma_2\sqrt{\frac{\ell_N}{N}}+ C \lambda_N\frac{\ell_N}{N}\\
    &+\frac{C_R N^{\gamma/2-1}}{\sqrt{\rho_N}} \bigg].
\end{split}
\end{equation}
The calibrated remainder term is $o(N^{-1/2})$ if
\begin{equation}
    N^{(\gamma-1)/2}\rho_N^{-1/2}\to0.
    \label{eq:growing-remainder-first-order}
\end{equation}
For $\rho_N=\delta_N$, this condition is equivalent to $\delta_N N^{1-\gamma}\to\infty$.

If $\ell_N$ is bounded or polylogarithmic, a sufficient power-law choice is $\lambda_N=N^\beta$ with
\begin{equation}
    \frac1{2(p-1)}<\beta<\frac12.
    \label{eq:growing-window}
\end{equation}
More generally, the threshold conditions are compatible whenever $\ell_N=o(N^{(p-2)/(2(p-1))})$.
\end{theorem}

\begin{proof}
The centered recursion is
\[
    y_{k+1}^{(N)} =(I-\eta_kH)y_k^{(N)}-\eta_kv_{N,k+1}.
\]
Its noise satisfies, uniformly in $N$,
\[
    \mathbb E v_{N,k}=0, \quad \mathbb E\|v_{N,k}\|^2 \le\mathbb E\|\clip_{\lambda_N}(\zeta)\|^2 \le\sigma_2^2.
\]
Also $\|b_N\|\le\mathbb E\|\zeta\|\le\sigma_2$, so the initial second moments of $y_1^{(N)}$ are uniformly bounded.

For all sufficiently large $k$, $\eta_k\le1/L$, and therefore
\[
    \mathbb E\|y_{k+1}^{(N)}\|^2 \le (1-\mu\eta_k)\mathbb E\|y_k^{(N)}\|^2 +\sigma_2^2\eta_k^2.
\]
Since $\gamma<1$, $\eta_k-\eta_{k+1}=o(\eta_k^2)$.
Choose $K$ so that, for $k\ge K$,
\[
    \eta_k-\eta_{k+1} \le\frac{\mu}{2}\eta_k^2.
\]
Taking a sufficiently large constant $C_y\ge2\sigma_2^2/\mu$ and applying induction gives
\[
    \mathbb E\|y_k^{(N)}\|^2\le C_y\eta_k,
\]
uniformly in $N$. The constant can be increased to cover the finite initial segment, whose second moments are also uniformly bounded. Proposition~\ref{prop:exact-pr} now yields \eqref{eq:growing-remainder-l2} and
\begin{equation}
    H\bar y_N^{(N)} =-\frac1N\sum_{k=1}^N v_{N,k+1}+R_N^{(N)}.
    \label{eq:growing-exact-pr-app}
\end{equation}

Because $\mathbb E\zeta=0$,
\[
    \|b_N\|\le\mathbb E\|\clip_{\lambda_N}(\zeta)-\zeta\|\le\frac{\mathbb E\|\zeta\|^p}{\lambda_N^{p-1}}.
\]
Hence
\begin{equation}
    \|e_N^\star\|\le\|H^{-1}\|\frac{\sigma_p^p}{\lambda_N^{p-1}}=o(N^{-1/2}).
    \label{eq:growing-root-shift-app}
\end{equation}

Since $\lambda_N\to\infty$, $\clip_{\lambda_N}(\zeta)\to\zeta$ in $L^2$. It follows that $\Cov(v_{N,1})\to\Sigma$. Moreover, $\|v_{N,k}\|\le2\lambda_N$ and $\lambda_N/\sqrt N\to0$ by \eqref{eq:growing-rigorous-conditions}. The triangular-array Lindeberg condition therefore holds, and
\[
    \frac1{\sqrt N}\sum_{k=1}^N v_{N,k+1}
    \Longrightarrow\mathcal N(0,\Sigma).
\]
Combine this limit with \eqref{eq:growing-exact-pr-app}, $\sqrt N R_N^{(N)}\to0$ in $L^2$, and \eqref{eq:growing-root-shift-app} to obtain \eqref{eq:growing-same-clt}.

For the finite-horizon bound, the Hilbert-space Bennett--Freedman inequality \cite{Pinelis1994} gives
\[
    \left\|\frac1N\sum_{k=1}^N v_{N,k+1}\right\| \le C\sigma_2\sqrt{\frac{\ell_N}{N}} +C\lambda_N\frac{\ell_N}{N}
\]
with probability at least $1-\delta_N$. Separately, Markov's inequality and
\eqref{eq:growing-remainder-l2} give
\[
    \mathbb P\left(\|R_N^{(N)}\| > \frac{C_R N^{\gamma/2-1}}{\sqrt{\rho_N}}\right)\le\rho_N.
\]
A union bound, followed by \eqref{eq:growing-exact-pr-app} and the bound on $e_N^\star$, proves \eqref{eq:growing-finite-bound}. No independence between these two events is required. Multiplying the remainder radius by $\sqrt N$ gives \eqref{eq:growing-remainder-first-order}.

Finally, for $\lambda_N=N^\beta$, the bias condition requires $\beta>1/[2(p-1)]$, while the bounded-increment condition holds for $\beta<1/2$ when $\ell_N$ is bounded or polylogarithmic. More generally, the lower scale $(\sigma_p^p\sqrt N)^{1/(p-1)}$ is asymptotically smaller than the upper scale $\sqrt N/\ell_N$ whenever $\ell_N=o(N^{(p-2)/(2(p-1))})$. Choosing a threshold strictly between these scales proves the last assertion.
\end{proof}

\begin{remark}[Scope of the threshold window]
The bias and bounded-increment requirements are sufficient conditions. At $p=2$, their lower and upper power exponents meet.

The theorem concerns linear additive noise and noise-centered clipping, with one threshold held constant throughout each horizon-$N$ run. Extensions to online thresholds, nonlinear recursions, or whole-gradient clipping require an additional asymptotic linearization and corresponding remainder control.
\end{remark}

\section{Additional Effects of Clipping and Constant Stepsizes}
\label{app:additional-results}

This appendix records the deterministic transit cost of whole-gradient
clipping and the constant-step covariance calculation for linear stochastic approximation.

\subsection{The deterministic transit cost of clipping}

The following exact calculation isolates the deterministic cost of clipping the optimization signal.

\begin{proposition}[linear transit versus geometric contraction]
\label{prop:transit}
Consider the noiseless one-dimensional quadratic $F(x)=\mu x^2/2$, $x_0>\lambda/\mu$, and a step $0<\eta\mu<1$.

For U-SGD, which is ordinary gradient descent here,
\begin{equation}\label{eq:unclip-transit}
    x_k=(1-\eta\mu)^k x_0, \quad T_{\rm U}(\lambda)= \left\lceil\frac{\log(\mu x_0/\lambda)}{-\log(1-\eta\mu)}\right\rceil
\end{equation}
steps suffice to reach $x_k\le\lambda/\mu$.  For whole-gradient clipping,
\begin{equation}
    x_{k+1}=x_k-\eta\lambda \quad\text{as long as } \; x_k>\lambda/\mu,
\end{equation}
so the transit time is
\begin{equation}\label{eq:clip-transit-time}
    T_{\rm C}(\lambda) = \left\lceil\frac{x_0 - (\lambda/\mu)}{\eta\lambda}\right\rceil.
\end{equation}
Thus a threshold selected from the noise scale can turn logarithmic deterministic contraction into a phase linear in $x_0/\lambda$.
\end{proposition}

Difference clipping can avoid this particular cost only after its anchor error, difference-moment certificate, refresh schedule, and auxiliary oracle requirements have been controlled.  Alternatively, a practical whole-gradient policy can use a large warm-up threshold and activate robust clipping only after localization.  Both modifications reduce the simplicity advantage of plain Clip-SGD and should be counted when algorithms are compared.

\phantomsection\label{Proof_prop:transit}
\begin{proof}
Without clipping, $x_{k+1}=(1-\eta\mu)x_k$, and solving $(1-\eta\mu)^kx_0\le\lambda/\mu$ gives \eqref{eq:unclip-transit}.  With clipping and $x_k>\lambda/\mu$, the direction equals $\lambda$, so $x_k=x_0-k\eta\lambda$ until the threshold region is reached.  Solving $x_0-k\eta\lambda\le\lambda/\mu$ gives \eqref{eq:clip-transit-time}.
\end{proof}

\subsection{The \texorpdfstring{$O(\eta)$}{O(eta)} covariance correction}

We now turn to the fine-grained constant-step regime of Mou et al.~\cite{MouEtAl2020}.  Consider linear stochastic approximation
\begin{equation}\label{eq:constant-lsa}
    e_{t+1}=\bigl(I-\eta A_{t+1}\bigr)e_t+\eta\varepsilon_{t+1}, \quad A_{t+1}=\bar A+\Xi_{t+1}.
\end{equation}

We use the following checkable conditions.  The pairs $(A_t,\varepsilon_t)$ are i.i.d. and independent of the past; $A_t$ and $\varepsilon_t$ are independent within each pair; $\E A_t=\bar A$, $\E\varepsilon_t=0$, and for some $q>4$, $\E\norm{A_t}_{\rm op}^q+\E\norm{\varepsilon_t}^q<\infty$. The matrix $\bar A$ is invertible with eigenvalues in the open right half-plane, and there exist $c>0$ and $\eta_0>0$ such that
\begin{equation}\label{eq:checkable-lsa-contraction}
    \left(\E\norm{I-\eta A_t}_{\rm op}^q\right)^{1/q} \le1-c\eta, \quad 0<\eta\le\eta_0.
\end{equation}

Condition \eqref{eq:checkable-lsa-contraction} implies a unique stationary solution with a finite $q$-th moment and geometric $L_q$ contraction.  We allow either stationary initialization or any $\cF_0$-measurable $e_0$, independent of the future i.i.d. pairs, with $\E\|e_0\|^q<\infty$; the contraction makes its initial transient negligible for the averaged CLT.  The condition is stronger than necessary, but avoids assuming ``a stationary solution and a CLT'' as hypotheses.

Let $\Sigma=\E\left[\varepsilon_t\varepsilon_t^\top\right]$ and let $\Lambda_\eta$ be the covariance of the stationary iterate.  Independence of the fresh pair from $e_t$ gives
\begin{equation}\label{eq:stationary-lyapunov}
    \bar A \Lambda_\eta+\Lambda_\eta\bar A^\top - \eta\bar A\Lambda_\eta\bar A^\top -\eta\E\left[\Xi\Lambda_\eta\Xi^\top\right] = \eta\Sigma.
\end{equation}

\begin{theorem}
\label{thm:mou-correction}
Under the preceding i.i.d., moment, independence, initialization, and contraction conditions, the recursion has a unique stationary law with covariance $\Lambda_\eta$.  For stationary initialization, and also for every independent finite-$q$ initialization described above, the averaged process satisfies a multivariate CLT with asymptotic covariance
\begin{equation}\label{eq:fine-covariance}
    \Gamma_\eta= \bar A^{-1} \left(\Sigma + \E\left[\Xi\Lambda_\eta\Xi^\top\right]\right) \bar A^{-\top}.
\end{equation}

If $\Lambda_0$ solves
\begin{equation}
    \bar A\Lambda_0 + \Lambda_0\bar A^\top = \Sigma,
\end{equation}
then, as $\eta\downarrow0$,
\begin{equation}\label{eq:fine-expansion}
    \Lambda_\eta = \eta\Lambda_0+O_F(\eta^2), \quad \Gamma_\eta=\bar A^{-1}\Sigma\bar A^{-\top} + \eta\bar A^{-1} \E\left[\Xi\Lambda_0\Xi^\top\right]\bar A^{-\top}  + O_F(\eta^2),
\end{equation}
where $O_F(\eta^2)$ is in Frobenius norm and is uniform for $0<\eta\le\eta_1$ for some $\eta_1\le\eta_0$.
\end{theorem}

The exact identity \eqref{eq:fine-covariance} and expansion \eqref{eq:fine-expansion} are specializations of the fixed-step i.i.d. LSA theory of Mou et al.~\cite{MouEtAl2020}.  Tight high-probability and moment bounds under weaker random-matrix conditions are developed by Durmus et al.~\cite{DurmusEtAl2021,DurmusEtAl2025}; quantitative Gaussian approximation and multiplier-bootstrap inference for decreasing-step and Markovian LSA are given in~\cite{samsonov2024gaussian,samsonov2026statistical}, with temporal-difference stability in~\cite{SamsonovEtAl2024TD}.  These results delimit the assumptions under which Gaussian concentration or inferential approximations are available.

For a fixed clipped score, freeze the local linearization at $x_\lambda$ and suppose that the resulting affine recursion satisfies the i.i.d. contraction conditions above. The formal substitution
\begin{equation}
    \bar A \rightsquigarrow A_\lambda, \quad \Sigma \rightsquigarrow \Sigma_\lambda, \quad \Xi \rightsquigarrow J_\lambda(\xi)-A_\lambda,
\end{equation}
where $J_\lambda(\xi)=\nabla_x\psi_\lambda(x_\lambda,\xi)$ away from the clipping sphere, identifies the \emph{multiplicative-noise component} of the local $O(\eta)$ correction. Taylor curvature terms, a generally nonzero $O(\eta)$ stationary-mean displacement, and possible score--Jacobian dependence contribute additional terms.  A full clipped nonlinear expansion would require a stationary bias theorem and a Poisson-equation/Markov-chain analysis uniform in the clipping parameter.  Step-size extrapolation results such as~\cite{HuoChenXie2026} also show that an $O(\eta)$ term is not an unimprovable risk contribution.

For the scalar illustration, take $A_t=1\pm0.6$ with equal probabilities and unit additive-noise variance.  Here
\begin{equation}
    \Lambda_\eta=\frac{\eta}{2-1.36\eta}, \quad \Gamma_\eta=1+0.36\Lambda_\eta =\frac{2-\eta}{2-1.36\eta}.
\end{equation}

The classical covariance is one, so this formula isolates the upward correction caused by a nonzero constant stepsize.

\subsubsection{Proofs for the covariance result}

\begin{lemma}[conditional covariance and Lindeberg for the effective LSA noise]
\label{lem:mou-effective-noise-clt}
Under the assumptions preceding Theorem~\ref{thm:mou-correction}, define $\Xi_t=A_t-\bar A$ and $m_{t+1}=\varepsilon_{t+1}-\Xi_{t+1}e_t$.  Let $\Lambda_\eta$ be the covariance under the unique stationary law.  For stationary initialization, and for every independent initial condition with finite $q$-th moment,
\begin{align}
    \frac1N \sum_{t=1}^N \E \left[m_{t+1}m_{t+1}^\top\mid\cF_t \right] &\xrightarrow{\Pp} \Sigma+ \E\left[\Xi\Lambda_\eta\Xi^\top \right], \label{eq:mou-conditional-covariance-app}
    \\ \frac1N\sum_{t=1}^N \E\!\left[\|m_{t+1}\|^2 \mathbf1 \left\{\|m_{t+1} \| > a\sqrt N \right\}\mid\cF_t\right]  & \xrightarrow{\Pp}0 \quad\text{for every }a>0. \label{eq:mou-conditional-lindeberg-app}
\end{align}
Consequently,
\begin{equation}\label{eq:mou-effective-noise-clt-app}
    \frac1{\sqrt N} \sum_{t=1}^Nm_{t+1} \quad \Longrightarrow \quad \mathcal N\! \left(0,\Sigma + \E\left[\Xi\Lambda_\eta\Xi^\top\right] \right).
\end{equation}
\end{lemma}

\begin{proof}
The fresh pair is independent of $\cF_t$, $\E\Xi=0$, and $\varepsilon$ is independent of $\Xi$. Hence $(m_{t+1},\cF_{t+1})$ is a martingale difference and, for the positive linear operator $\mathcal T(B)=\E[\Xi B\Xi^\top]$,
\begin{equation}
    \E[m_{t+1}m_{t+1}^\top\mid\cF_t] =\Sigma + \mathcal T(e_te_t^\top).
\end{equation}
Under stationary initialization, geometric $L_q$ contraction implies ergodicity, so the ergodic theorem applied entrywise to $\mathcal T(e_te_t^\top)$ gives \eqref{eq:mou-conditional-covariance-app}.  For a nonstationary finite-$q$ initial condition, couple $e_t$ with a stationary copy $\widetilde e_t$ driven by the same fresh pairs.  Iterating \eqref{eq:checkable-lsa-contraction} gives
\begin{equation}
    \{\E\|e_t-\widetilde e_t\|^q\}^{1/q} \le (1-c\eta)^t \{\E\|e_0-\widetilde e_0\|^q\}^{1/q}.
\end{equation}
Because $q>4$, Hölder's inequality, the finite $q$-moments of $\Xi,e_t$ and $\widetilde e_t$, and $\|uu^\top-vv^\top\|_F\le(\|u\|+\|v\|)\|u-v\|$ show that the Cesaro mean of $\|\mathcal T(e_te_t^\top)-\mathcal T(\widetilde e_t\widetilde e_t^\top)\|_F$ converges to zero in $L^1$. Thus the same covariance limit holds from every allowed initialization.

For Lindeberg, conditional Markov's inequality gives
\begin{align}
    \frac1N \sum_{t=1}^N \E\!\left[\|m_{t+1}\|^2 \mathbf1 \left\{\|m_{t+1}\|>a \sqrt N\right\} \mid\cF_t\right] & \le \frac{1}{a^{q-2}N^{(q-2)/2}} \frac1N \sum_{t=1}^N\E[\|m_{t+1}\|^q\mid\cF_t] 
    \\& \le \frac{C}{N^{(q-2)/2}} \left(1+\frac1N\sum_{t=1}^N\|e_t\|^q\right). 
\end{align}

The last empirical average is $O_{\Pp}(1)$ by the stationary ergodic theorem and the same coupling argument. Since $q>4$, the right-hand side converges to zero. The multivariate martingale CLT applied with \eqref{eq:mou-conditional-covariance-app}--\eqref{eq:mou-conditional-lindeberg-app} proves \eqref{eq:mou-effective-noise-clt-app}.
\end{proof}

\begin{proof}[Proof of Theorem~\ref{thm:mou-correction}]
In stationarity, $e_t$ is independent of the fresh pair $(A_{t+1},\varepsilon_{t+1})$.  Its mean $m_\eta$ satisfies $m_\eta=(I-\eta\bar A)m_\eta$; invertibility of $\bar A$ therefore gives $m_\eta=0$. Expanding the covariance of \eqref{eq:constant-lsa} gives
\begin{align}
    \Lambda_\eta &= \E[(I-\eta A)\Lambda_\eta(I-\eta A)^\top] +\eta^2\Sigma
    \\& = \Lambda_\eta -\eta(\bar A\Lambda_\eta+\Lambda_\eta\bar A^\top) +\eta^2\bar A\Lambda_\eta\bar A^\top +\eta^2\E\left[\Xi\Lambda_\eta\Xi^\top\right] +\eta^2\Sigma.
\end{align}
Rearrangement proves \eqref{eq:stationary-lyapunov}.

The recursion also gives the exact identity
\begin{equation}\label{eq:constant-step-average-identity}
    \bar A e_t = \varepsilon_{t+1}-\Xi_{t+1} e_t -\frac{e_{t+1}-e_t}{\eta}.
\end{equation}
The effective noise
\begin{equation}
    m_{t+1}=\varepsilon_{t+1}-\Xi_{t+1}e_t
\end{equation}
has the martingale CLT established in Lemma~\ref{lem:mou-effective-noise-clt}. Summing \eqref{eq:constant-step-average-identity} gives
\begin{equation}
    \bar A\sqrt N\,\bar e_N =\frac1{\sqrt N}\sum_{t=1}^Nm_{t+1} -\frac{e_{N+1}-e_1}{\eta\sqrt N}.
\end{equation}
The contraction assumption gives $\sup_t\E\|e_t\|^q<\infty$, so the boundary term is $o_{\Pp}(1)$ for every allowed initialization. Lemma~\ref{lem:mou-effective-noise-clt} and Slutsky's theorem yield \eqref{eq:fine-covariance}.

Write $\Lambda_\eta=\eta\Lambda_0+R_\eta$ in \eqref{eq:stationary-lyapunov}. Invertibility of the continuous Lyapunov operator $X\mapsto\bar AX+X\bar A^\top$ gives $\norm{R_\eta}=O(\eta^2)$.  Substitution into \eqref{eq:fine-covariance} proves \eqref{eq:fine-expansion}.
\end{proof}

\section{Experimental Protocol}
\label{app:experiments}
\label{app:obd-protocol}

This appendix gives the protocol for the OBD experiment in
Section~\ref{sec:experiments}. The source and target empirical risks,
the compared directions, and the output average are defined there.

\subsection{Source-only selection and paired evaluation}

We freeze the experimental protocol before target evaluation. First, one U-SGD stepsize is selected on $128$ source-objective validation trajectories and then shared by every method, so clipping is the only algorithmic change. Next, a second set of \OBDThresholdValidationRuns{} trajectories selects one threshold from the predeclared grid $\{0.05,0.1,0.2,0.5,1,2,5,10,50,200,\infty\}$ by minimizing the source $0.95$-quantile at $N=\OBDThresholdValidationHorizon{}$; exact ties are resolved toward the largest threshold. This selects $\lambda_{\rm val}= \OBDSelectedThreshold$, where the subscript emphasizes that this is the validation-selected value of the same clipping threshold $\lambda$. The Random-policy log is not used in either selection stage. We then freeze $\lambda_{\rm val}$ and evaluate it against U-SGD on \OBDEvaluationRuns{} fresh trajectories, using common sampled rows between methods. The complete grid is evaluated separately as a threshold-sensitivity diagnostic.

\subsection{Uncertainty and threshold sensitivity}

Figure~\ref{fig:obd-tradeoff} reports pointwise distribution-free
$95\%$ order-statistic intervals conditional on the fixed logs.
The paired-bootstrap intervals in the main text quantify the reported
clipped-to-unclipped error ratios. The threshold sweep is a diagnostic
over the predeclared grid and is not used to tune the frozen method.

\subsection{Tail diagnostic}

Figure~\ref{fig:obd-tradeoff}(a) characterizes the empirical tails of the importance weights and stochastic gradients. At horizontal value $x$, its vertical value is the fraction of observations whose weight or gradient norm exceeds $x$ times its median. The long right tails show that a tiny fraction of observations is orders of magnitude larger than a typical one: observations at or above the empirical $0.999$-quantile of the gradient norm account for $99.6\%$ of its empirical second moment. This concentration of the empirical second moment motivates the clipping comparison.

\section{Directions for Accelerated Methods}
\label{sec:acceleration}

Accelerated clipped methods under heavy-tailed noise already admit high-probability guarantees~\cite{GorbunovDanilovaGasnikov2020}, while unmodified accelerated proximal schemes can be optimal in expectation under weak moments~\cite{HeLu2025,LiuExpectation2026}.  A direct counterpart of our three-level comparison is still missing.  We therefore record only the weighted quantities that such a theorem would have to control.

\subsection{Weighted Nagaev geometry}

Suppose an accelerated output admits a predictable stochastic representation
\begin{equation}\label{eq:accelerated-weighted-noise}
    Z_N=\sum_{k=1}^N w_{k,N}\zeta_k, \quad w_{k,N}\ge0, \quad \sum_{k=1}^Nw_{k,N}=1.
\end{equation}

Corollary~\ref{cor:weighted-nagaev} then gives
\begin{equation}\label{eq:accelerated-nagaev}
    \norm{Z_N} \lesssim \sigma_2\sqrt{\ell_\delta V_{2,N}} +\sigma_p\delta^{-1/p}V_{p,N}^{1/p}, \quad V_{r,N}=\sum_{k=1}^Nw_{k,N}^r.
\end{equation}

Define
\begin{equation}
    N_2=V_{2,N}^{-1}, \quad N_p=V_{p,N}^{-1/(p-1)}, \quad w_{\max,N}=\max_k w_{k,N}.
\end{equation}

Uniform averaging has $N_2=N_p=N$. Momentum can concentrate the weights, reducing $N_p$ and increasing the maximum influence. The resulting candidate Gaussian-dominance boundary is
\begin{equation}\label{eq:accelerated-crossover}
     \delta\ell_\delta^{p/2} \gtrsim \chi_p^p\frac{N_2^{p/2}}{N_p^{p-1}}.
\end{equation}

\begin{hypothesis}[accelerated regime map]
\label{hyp:accelerated-crossover}
If a concrete accelerated convex method has deterministic error $O(LR_0^2/N^2)$ and a localized predictable representation of the form \eqref{eq:accelerated-weighted-noise}, then its candidate heavy-tail stochastic term is \eqref{eq:accelerated-nagaev}.  For $p>2$, U-SGD-style untrimmed acceleration should match the leading clipped order in the region \eqref{eq:accelerated-crossover}; outside it, clipping can improve confidence dependence but may attenuate the accelerated signal.
\end{hypothesis}
The hard step is the predictable representation after momentum, clipping, and localization have interacted.

\subsection{Thresholds, centering, and first-order efficiency}

For a common threshold, the bias term is of order $\sigma_p^p/\lambda^{p-1}$ and the bounded-increment contribution is of order $\lambda\ell_\delta w_{\max,N}$.  Balancing them suggests
\begin{equation}
    \lambda_\star \asymp \sigma_p(w_{\max,N}\ell_\delta)^{-1/p}, \quad \text{correction } \asymp \sigma_p(w_{\max,N}\ell_\delta)^{1-1/p}.
\end{equation}

A sufficient weighted analogue of the robust--efficient window is
\begin{equation}\label{eq:weighted-window}
    \left(\frac{\sigma_p^p}{\sigma_2\sqrt{V_{2,N}}}\right)^{1/(p-1)} \ll\lambda_N \ll \frac{\sigma_2\sqrt{V_{2,N}}}{\ell_Nw_{\max,N}}.
\end{equation}

For uniform weights this reduces to \eqref{eq:growing-window}; concentrated accelerated weights can shrink or eliminate the interval.  An implementable difference-clipped accelerated method would additionally need a charged anchor estimator and a proof that one clipped error remains controlled through all later momentum states.

Under strong convexity, a successful construction should retain the deterministic $\sqrt\kappa\log(\Delta_0/\eps)$ burn-in while preserving the leading variance term $\sigma_2^2/(\mu\eps)$.  For fine-grained limits, the augmented position--momentum state leads to a block Lyapunov operator.  A proof would need exactly four ingredients: the stochastic weights, a localized weighted Nagaev decomposition, a maximum-influence bound through the augmented dynamics, and a triangular-array linearization uniform in the threshold.

\end{document}